\documentclass{article}

\usepackage{basic_arxiv}

\usepackage[utf8]{inputenc}      % allow utf-8 input
\usepackage[T1]{fontenc}         % use 8-bit T1 fonts
\usepackage[hidelinks]{hyperref} % hyperlinks
\usepackage{url}                 % simple URL typesetting
\usepackage{booktabs}            % professional-quality tables
\usepackage{nicefrac}            % compact symbols for 1/2, etc.
\usepackage{microtype}           % microtypography
\usepackage{lipsum}
\usepackage{graphicx}
\usepackage[numbers]{natbib}     % Or [authoryear] for author-year citations
\usepackage{caption}             % For making caption text smaller
\title{The dynamic interplay between in-context and in-weight learning in humans and neural networks}

\author{
  Jacob Russin \\
  Department of Computer Science\\
  Department of Cognitive and Psychological Sciences\\
  Brown University\\
  \href{mailto:jake_russin@brown.edu}{jake\_russin@brown.edu}\\
\AND
  Ellie Pavlick\thanks{E.P. and M.J.F. are joint senior authors on this work.}\\
  Department of Computer Science\\
  Brown University \\
\And
  Michael J. Frank\footnotemark[1]\\
  Department of Cognitive and Psychological Sciences\\
  Carney Institute for Brain Science\\
  Brown University\\
}

\begin{document}
\maketitle
\setcounter{footnote}{0} % Starting at 0 again after putting the \thanks above
\begin{abstract} % Currently 230 words
Human learning embodies a striking duality: sometimes, we appear capable of following logical, compositional rules and benefit from structured curricula (e.g., in formal education), while other times, we rely on an incremental approach or trial-and-error, learning better from curricula that are randomly interleaved.
Influential psychological theories explain this seemingly disparate behavioral evidence by positing two qualitatively different learning systems---one for rapid, rule-based inferences and another for slow, incremental adaptation. 
It remains unclear how to reconcile such theories with neural networks, which learn via incremental weight updates and are thus a natural model for the latter type of learning, but are not obviously compatible with the former. 
However, recent evidence suggests that metalearning neural networks and large language models are capable of ``in-context learning'' (ICL)---the ability to flexibly grasp the structure of a new task from a few examples.
Here, we show that the dynamic interplay between ICL and default in-weight learning (IWL) naturally captures a broad range of learning phenomena observed in humans, reproducing  curriculum effects on category-learning and compositional tasks, and recapitulating a tradeoff between flexibility and retention.
Our work shows how emergent ICL can equip neural networks with fundamentally different learning properties that can coexist with their native IWL, thus offering a novel perspective on dual-process theories and human cognitive flexibility.
\end{abstract}

\section*{Introduction}
Humans are capable of two qualitatively distinct kinds of learning \cite{AshbyMaddox11, BotvinickRitterWangEtAl19a, DawNivDayan05, Evans08, NohYanBjorkEtAl16, OReillyNairRussinEtAl20, Sable-MeyerBenjaminWatkinsEtAl24, Sloman96a, CollinsFrank18b}.
The first involves slow, incremental adaptation to the environment through trial and error \cite{BotvinickRitterWangEtAl19a, CollinsFrank18b, OReillyNairRussinEtAl20}.
The second is much more advanced and involves rapid inference of rules or structure from information available in the environment or held in working memory \cite[WM;][]{CollinsFrank13a, CollinsFrank16b, LakeUllmanTenenbaumEtAl17a, OReillyFrank06, RougierNoelleBraverEtAl05d}.
For example, although it can famously take 10,000 hours to master the violin, when given a mandolin for the first time an expert musician may rapidly infer the rules about how each string is tuned.

Many findings support the idea that humans exhibit different learning and generalization behaviors in different domains \cite{AshbyMaddox11, FleschBalaguerDekkerEtAl18, NohYanBjorkEtAl16, PesnotLerousseauSummerfield24, Rac-LubashevskyCremerCollinsEtAl23, CollinsFrank16b, CollinsFrank18b}.
On the one hand, in tasks that are readily described by simple rules (e.g., where only one stimulus feature is relevant), humans learn efficiently from only a few examples (``few-shot learning''), appearing to make rapid inferences about the latent structure governing the task \cite{LakeUllmanTenenbaumEtAl17a, LakeSalakhutdinovGrossEtAl11, LakeSalakhutdinovTenenbaum15b}. 
When this latent structure is \emph{compositional}, humans can generalize by flexibly recombining familiar elements according to inferred rules \cite{DekkerOttoSummerfield22c, FodorPylyshyn88a, FranklinFrank18a, FranklinFrank20a, FranklandGreene20, LakeLinzenBaroni19a, LiuFrank22, RussinEtAl2024, SchwartenbeckBaramLiuEtAl23}. 
In such settings, people exhibit a \emph{blocking advantage}, learning better when information is organized into blocks of related examples that make this underlying structure more salient \cite{BeukersCollinKempnerEtAl24a, DekkerOttoSummerfield22c, FleschBalaguerDekkerEtAl18, NohYanBjorkEtAl16}.
On the other hand, when a task is not governed by simple rules, learning may require integrating across multiple task dimensions. 
This kind of learning proceeds more incrementally \cite{AshbyMaddox11, FrankBadre12, NohYanBjorkEtAl16}, but is also associated with greater \textit{retention} after a delay \cite{CollinsFrank18b, Rac-LubashevskyCremerCollinsEtAl23}. 
In these contexts, compositional generalization is not possible, and, as shown in both laboratory \cite{NohYanBjorkEtAl16, RichlandFinleyBjork04} and real-world contexts \cite{GoodeMagill86, LandinHebertFairweather93}, people exhibit an \emph{interleaving advantage}, learning better when trials are randomly shuffled over time. 

Dual-process accounts \cite{AshbyMaddox11, Evans08, Kahneman11, NohYanBjorkEtAl16}, explain these contrasting effects by positing two separate learning systems: a rule-based or symbolic system that is compositional and operates by testing explicit hypotheses, and a procedural or sub-symbolic system that learns more incrementally and can capture arbitrary associations, even in the absence of simple rules.
Neural networks, which learn via incremental weight updates \cite{RumelhartHintonWilliams86, RumelhartMcClellandPDPResearchGroup86}, offer a natural framework for understanding the latter.
However, it is less clear how they could explain the features associated with the rule-based system, namely, flexible (few-shot) learning, compositionality, and the blocking advantage. 
Neural networks are notoriously data hungry compared to human learners \cite{LakeUllmanTenenbaumEtAl17a}, and have been criticized for failing to account for human compositionality \cite{FodorPylyshyn88a, LakeLinzenBaroni19a, LakeBaroni18, Marcus98, Marcus18}, as they do not explicitly represent rules or symbols \cite{PinkerPrince88a, LakeUllmanTenenbaumEtAl17a, Quilty-DunnPorotMandelbaum23}.
Furthermore, the blocked curricula that confer an advantage to humans on some tasks \cite{FleschBalaguerDekkerEtAl18, NohYanBjorkEtAl16} generally result in ``catastrophic forgetting'' in neural networks, because new learning can overwrite information stored in the same weights during previous blocks \cite{RussinZolfagharParkEtAl22, McCloskeyCohen89, McClellandMcNaughtonOReilly95}.

Some biologically informed models account for various aspects of dual-process theories \cite[e.g., ][]{LoveMedinGureckis04a}, such as models of prefrontal cortex (PFC) that emphasize the importance of dynamic activation-based representations for inferring rules and flexibly adapting to the current context \cite{CollinsFrank13a, FrankClaus06, KrieteNoelleCohenEtAl13, MillerCohen01, RougierNoelleBraverEtAl05d, WangKurth-NelsonKumaranEtAl18a}. 
However, these models have not confronted how the emergence of such rule-based processing might relate to curriculum effects and compositionality.

\begin{figure*}
\centering
\includegraphics[width=.75\linewidth]{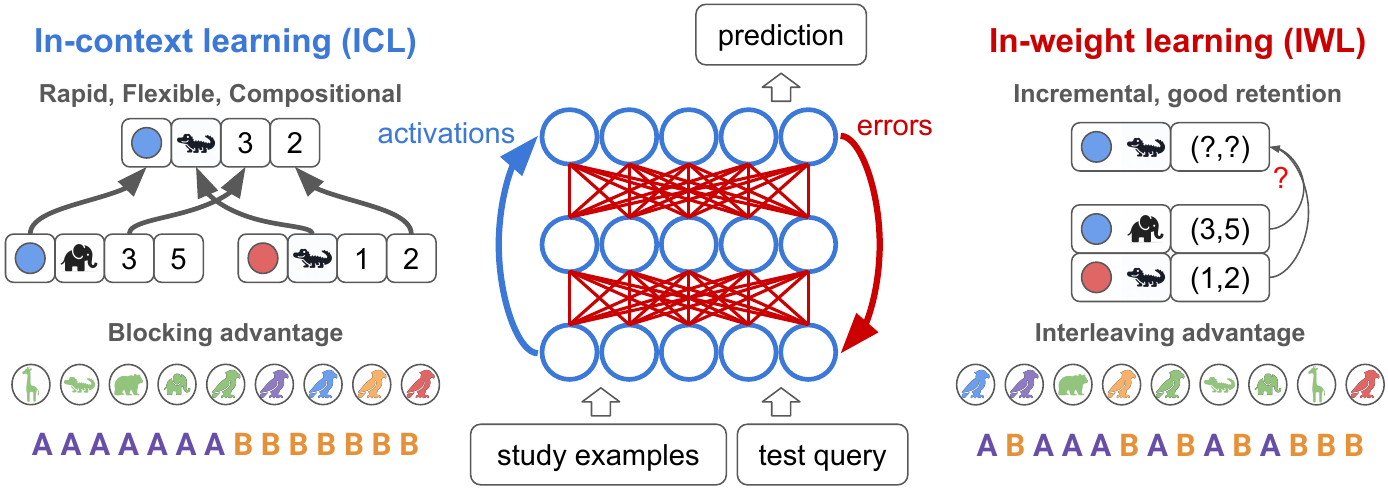}
\caption{Properties of in-context learning (ICL) and in-weight learning (IWL). ICL (blue) is the ability of a neural network to flexibly learn a new task from just a few study examples given in context, and to apply this knowledge to a novel test query (also given in context). ICL is carried out through the activation dynamics of the network (blue arrow), and can happen without weight updates. ICL can be flexible and compositional, and is shown here predicting the location of a blue alligator (x=3, y=2) by composing elements of the known locations of a blue elephant (x=3) and a red alligator (y = 2). ICL can exhibit a blocking advantage, learning better when related examples are blocked over time. IWL (red) is the usual form of learning in neural networks, wherein prediction errors are backpropagated to update weights. IWL can result in better retention but is non-compositional, depicted here as failing to generalize to the blue alligator due to its reliance on a simple lookup table that ignores the compositional structure of the task. IWL exhibits an interleaving advantage, learning better when examples are randomly shuffled or interleaved due to the well-known problem of catastrophic forgetting.}
\label{fig:intro}
\end{figure*}

We hypothesized that neural networks with greater cognitive flexibility can simultaneously reproduce human-like behaviors related to few-shot generalization, compositionality and curriculum effects. 
Recent advances in artificial neural networks have demonstrated surprising success on rule-governed tasks involving reasoning \cite{BrownMannRyderEtAl20b, BubeckChandrasekaranEldanEtAl23, SaparovPangPadmakumarEtAl23}, analogy \cite{MuskerDuchnowskiMilliereEtAl24, WebbHolyoakLu23}, and compositionality \cite{LakeBaroni23a, PressZhangMinEtAl23a, ZhouScharliHouEtAl22a}.
Many of these capabilities are connected to the emergence of \emph{in-context learning} (ICL), or the ability to learn new tasks from demonstrations or instructions given in context \cite{BrownMannRyderEtAl20b, ChanSantoroLampinenEtAl22, vonOswaldNiklassonSchlegelEtAl23, XieRaghunathanLiangEtAl22}.
For example, if demonstrations of a novel task are provided as contextual inputs (strawberry → red, banana → yellow), trained networks such as large language models (LLMs) can often readily perform the task on new inputs (plum → ??).

Importantly, ICL does not require updates to network weights.\footnote{One may prefer to think of ICL not as a kind of ``learning,'' but rather as a kind of \textit{inference}. This terminological issue can also arise in the case of human few-shot learning. We have chosen to adopt the ``learning'' terminology as it has become standard in machine learning research.} 
This stands in contrast with \emph{in-weight learning} (IWL)---the usual form of learning in neural networks---which proceeds by backpropagating errors to update weights \cite{RumelhartHintonWilliams86}.
Instead, ICL takes place within the model’s activation dynamics, which similarly support cognitive flexibility and working memory (WM) in neural network models of PFC \cite{OReillyFrank06, RougierNoelleBraverEtAl05d, CollinsFrank13a, WangKurth-NelsonSoyerEtAl17}. 
Thus, the emergence of ICL results in a tradeoff \cite{AnandLeporiMerulloEtAl24, ChanSantoroLampinenEtAl22, Reddy23a}: when ICL succeeds, fewer errors are accumulated, resulting in fewer weight updates. 
We reasoned that such a tradeoff could account for a variety of phenomena in human learning. 
For example, when information can be learned rapidly within WM, neural prediction errors are suppressed \cite{CollinsFrank18b, CollinsCiulloFrankEtAl17, CollinsFrank16b}. 
This neural signature predicts enhanced generalization of rule-like structure \cite{CollinsFrank16b} but {\em degraded} reinforcement learning and less robust retention \cite{Rac-LubashevskyCremerCollinsEtAl23, HitchcockKimFrank24}.  

Advanced ICL abilities have been shown to emerge in LLMs \cite{BrownMannRyderEtAl20b, BubeckChandrasekaranEldanEtAl23}, but can also be imparted more directly via metalearning, where a network is specifically trained (via the usual form of IWL) to \emph{learn how to learn} new tasks provided in context \cite{BinzDasguptaJagadishEtAl23a, FinnAbbeelLevine17d, JagadishCoda-FornoThalmannEtAl24, SantoroBartunovBotvinickEtAl16, WangKurth-NelsonSoyerEtAl17, WangKurth-NelsonKumaranEtAl18a}.
Metalearning networks that perform ICL through their activation dynamics have been shown to reproduce phenomena associated with the PFC \cite{RougierNoelleBraverEtAl05d, CollinsFrank13a, WangKurth-NelsonKumaranEtAl18a}, and human-like compositional generalizations \cite{LakeBaroni23a}, suggesting that emergent ICL abilities can result in more rule-like or compositional behaviors than the standard IWL used to train networks in the first place \cite{RussinEtAl2024, RussinMcGrathEtAl2024}.

In this work, we demonstrate how a single neural network capable of both ICL and IWL can simultaneously replicate the behavioral effects associated with each of the two systems posited in traditional dual-process theories \cite{AshbyMaddox11, Evans08, NohYanBjorkEtAl16}, producing compositional generalizations and the blocking advantage in rule-governed tasks, while exhibiting an interleaving advantage in tasks lacking such structure. 
Moreover, we show how the very same mechanisms can give rise to the tradeoff between flexibility and retention observed in human reinforcement learning \cite{Rac-LubashevskyCremerCollinsEtAl23, CollinsFrank16b,HitchcockKimFrank24}.

Our theoretical framework can be summarized by three key principles (see Figure~\ref{fig:intro}):
\begin{enumerate}
    \setlength{\itemsep}{0.1em} % controls space between items
    \setlength{\parskip}{0pt} % controls space between paragraphs within an item
    \item \label{item:IWL} \textbf{Properties of IWL}: IWL fails on compositional generalization problems, shows an interleaving advantage due to catastrophic forgetting when trials are blocked, and results in better retention.
    \item \label{item:ICL} \textbf{Properties of ICL}: ICL can be endowed with inductive biases that produce compositional generalization and a blocking advantage, but results in worse retention.
    \item \label{item:both} \textbf{Dynamic interplay}: when ICL is possible, its properties dominate because few errors are made, suppressing IWL. But when ICL is difficult, the properties of IWL dominate because errors result in larger weight updates.
\end{enumerate}
We test this theoretical framework by experimenting with metalearning neural networks on tasks used in previous human studies \cite{NohYanBjorkEtAl16, DekkerOttoSummerfield22c}, showing how the dynamic interplay between ICL and IWL offers a unified account of learning phenomena observed in humans across a wide range of studies from cognitive psychology and neuroscience\footnote{All code used for simulations is available at https://github.com/jlrussin/icl-iwl-interplay.}. 
First, we show in a category-learning setting \cite{NohYanBjorkEtAl16} that a single neural network capable of ICL and IWL produces both of the curriculum effects observed in humans---a blocking advantage in the presence of rule-like structure, and an interleaving advantage in the absence of such structure.
Then, we show that when applied to a compositional task, the neural network produces the compositional generalization behaviors and the blocking advantage observed in humans on the same task \cite{DekkerOttoSummerfield22c}. 
We then test existing LLMs on this compositional task and show that emergent ICL in these models exhibits both compositionality and a blocking advantage as well.
Finally, we demonstrate how the dynamic interplay between ICL and IWL naturally gives rise to a tradeoff between flexibility and retention observed in recent human studies \cite{CollinsFrank18b, Rac-LubashevskyCremerCollinsEtAl23, HitchcockKimFrank24}.

The primary goal of this work is not simply to show that neural networks can perform well on these cognitive tasks, but to demonstrate how the specific principles governing the dynamic interplay between ICL and IWL naturally reproduce these particular learning phenomena.
Our findings show how these two qualitatively distinct learning processes can interact within a single neural network model, and offer a novel perspective on how dual-process theories of cognition might be reconciled with a neural network perspective.

\section*{Results} 
\subsection*{Curriculum effects in category-learning} 

We first consider whether the principles above can account for the curriculum effects observed in human category learning, before turning to compositionality in the next section.
As reviewed above, humans exhibit an interaction between the task (rule-like vs.\ rotated), and curriculum (blocked vs.\ interleaved) in category learning ($\eta_p = 0.04$), showing a blocking advantage ($d=0.47$) when categories are governed by succinct rules, but an interleaving advantage ($d=0.33$) when no such rules are readily available \cite{NohYanBjorkEtAl16}.

\begin{figure*}%[tbhp]
\centering
\includegraphics[width=.99\linewidth]{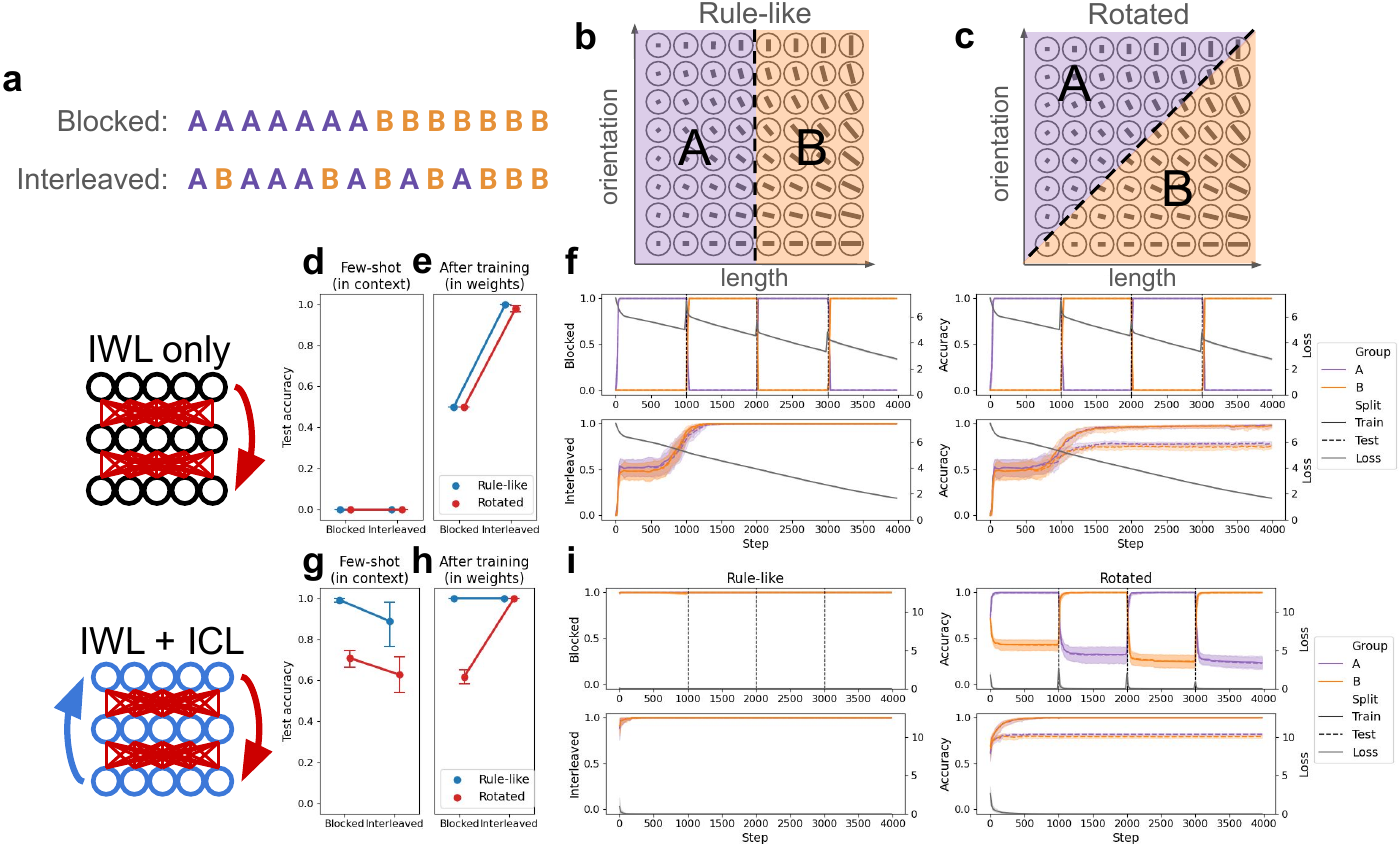}
\caption{Category-learning experiments. The task is derived from a human study \cite{NohYanBjorkEtAl16}. Networks were presented with multi-feature items along with their category labels, and tested on unseen items. See Appendix for details. \textbf{(a)} Curriculum conditions. Trials were either blocked by category label or randomly interleaved. \textbf{(b)} In the rule-like condition, category membership was determined by a simple rule that only depended on one of the two features (e.g., `A' if length $\leq$ 4, `B' otherwise). \textbf{(c)} In the rotated (information-integration) condition, category membership was jointly determined by both feature dimensions. The original axes were rotated by 45 degrees and a category boundary was chosen in the new coordinate system. \textbf{(d-f)} Category-learning with in-weight learning (IWL) only. Randomly initialized networks were trained from scratch on the task. \textbf{(d)} The few-shot evaluation tested networks' ability to learn the task from the 32 examples presented in context, before any weight updates were made. Unsurprisingly, randomly initialized networks without prior metalearning experience were incapable of utilizing the examples given in context to learn the task, regardless of condition. Values correspond to the average test accuracies shown in (f), but before step 0 (i.e., before any training). Note that model choices were not forced to be one of the two category labels, so chance performance here corresponds to $1/d_v$, where $d_v$ is the vocabulary size. \textbf{(e)} The same evaluation was conducted after training, testing networks' ability to learn through IWL. Without prior metalearning experience, the network was able to learn in-weights, performing well on both the rule-like and the rotated tasks after training. However, performance was much worse in the blocked condition due to catastrophic forgetting (see f). Here, values correspond to the train accuracy (i.e., accuracy on the 32 train items) in (f) at the final timestep. \textbf{(f)} Accuracy and loss results over the course of IWL training in each of the four conditions. Accuracy is split by category to better visualize the effects of catastrophic forgetting in the blocked condition (top row). \textbf{(g-i)} Category learning with both in-weight and in-context learning (ICL). Networks first metalearned on a distribution of related tasks (not shown), and were subsequently trained on specific category-learning tasks from each condition. \textbf{(g)} After metalearning, the models exhibited strong ICL on the task, as shown by the high few-shot accuracy. ICL demonstrated a blocking advantage, and also showed improved performance in the rule-like compared to the rotated condition. \textbf{(h)} After training had occurred on a specific task, the network exhibited an interleaving advantage in the rotated condition, due to catastrophic forgetting when trials were blocked (see i). \textbf{(i)} Accuracy and loss results over the course of task-specific training in each of the four conditions. When trials were blocked in the rule-like condition, ICL achieved near-perfect accuracy immediately, resulting in little loss and thus little IWL. When trials were interleaved, few-shot performance was worse (see g), but performance quickly recovered due to compensation by IWL. In the rotated condition, ICL failed, resulting in larger losses and increased IWL. This IWL resulted in catastrophic forgetting, as can be seen in the rapid decline in performance on `A' items while training on `B', and vice versa. No such catastrophic forgetting occurred when trials were interleaved (although test performance was not perfect).}
\label{fig:category}
\end{figure*}

We designed a category-learning task directly based on this previous work \cite{NohYanBjorkEtAl16}, but suitable for use with metalearning neural networks (see Figure~\ref{fig:category}a-c).
Stimuli varied along two feature dimensions (akin to line length and line orientation) with 8 possible values, yielding 64 possible items.
Each item was assigned to one of two categories, indicated by an arbitrary category label (e.g., `A' or `B').
In the \textbf{Rule-like} (or ``Rule-based'') condition, one of the two feature dimensions determined category membership (e.g., lines with shorter lengths are in category `A' and lines with longer lengths are in category `B'), while in the \textbf{Rotated} (or ``information-integration'') condition, category membership was determined by both features. 
This simple rotation has been shown to challenge the search for a simple, verbalizable rule, and is thought to recruit the more incremental procedural learning system in humans \cite{NohYanBjorkEtAl16, AshbyMaddox11}.
Networks were presented with 16 items from each category (32 total), and tested on the remaining held-out items.
The 32 items used during learning were either \textbf{Blocked}, where items from one category were presented first, followed by the items from the other, or \textbf{Interleaved}, where items were randomly shuffled.
Both rotation conditions were tested with both curriculum conditions, yielding a 2x2 design.

\subsubsection*{IWL produces an interleaving advantage} 
In this category-learning setting, a network capable of IWL but not ICL exhibited an interleaving advantage, regardless of the presence or absence of rule-like structure.
This is consistent with classic findings showing that standard learning in neural networks (i.e., IWL) benefits from random interleaving due to the well-known phenomenon of catastrophic forgetting \cite{McCloskeyCohen89, McClellandMcNaughtonOReilly95}.
A randomly initialized network was trained from scratch on the categorization task in each of the four conditions (see methods for details).
Because IWL requires slow, incremental updates, this network was not capable of few-shot learning in this setting (see Figure~\ref{fig:category}d) even in the rule-like condition, where a few examples should suffice for inference of the simple rule.
Consistent with our theoretical framework (principle~\ref{item:IWL}), the model performed better when trials were interleaved compared to when they were blocked ($p < 10^{-3}$; see Figure~\ref{fig:category}e-f), in both the rule-like and rotated conditions (although slightly better in the rule-like condition).
This interleaving advantage was due to catastrophic forgetting when trials were blocked, which can be seen in the dramatic decrease in the network’s performance on examples of the category trained during the previous block (e.g., performance on category `A' decreases as category `B' is trained in the second block).
Thus, the default in-weight learning (IWL) behavior of neural networks can explain why an interleaving advantage would be observed in human category-learning \cite{NohYanBjorkEtAl16}. 
However, a network capable of IWL alone cannot account for the blocking advantage that humans exhibit when categories are governed by rule-like structure \cite{DekkerOttoSummerfield22c, FleschBalaguerDekkerEtAl18, NohYanBjorkEtAl16}.

\subsubsection*{ICL can produce a blocking advantage} 
Next, we endowed a network with ICL abilities by having it metalearn on a distribution of categorization tasks (see methods for details). 
Metalearning can induce ICL in deep neural networks \cite{LakeBaroni23a, vonOswaldNiklassonSchlegelEtAl23}, and relatedly, has been shown to give rise to abstract generalizable representations in models of PFC \cite{RougierNoelleBraverEtAl05d, WangKurth-NelsonKumaranEtAl18a, CollinsFrank13a}.
These ICL abilities allowed the network to solve unseen tasks given in context through its activation dynamics, even when weights were frozen and no IWL was allowed to occur. 

To ensure that ICL would have the desired properties (see principle~\ref{item:ICL}), we had the network metalearn on a distribution of categorization tasks with 1) rule-like structure and 2) blocked curricula.
We then evaluated the network in the few-shot setting, where the weights were frozen and the network had to learn new tasks from a few examples given in context (see methods for details; see Figure~\ref{fig:category}g).
As predicted, when the model had developed ICL abilities by metalearning on rule-like category-learning problems, it could generalize to new rule-like problems, but struggled to solve tasks in context in the rotated condition (main effect of rotation: $p < 10^{-3}$).
Moreover, ICL exhibited a blocking advantage on unseen rule-like categorization problems (main effect of curriculum: $p < 0.05$).
This blocking advantage also emerged due to the metalearning distribution (see Appendix), but see Discussion for alternative explanations based on architectural constraints in human brains. 
In sum, these few-shot results suggest that it is possible to endow a network with ICL abilities that are sensitive to rule-like structure and learning curriculum: the network was capable of making inferences over the items provided in context, but was better at doing so when related items were organized into blocks.

\subsubsection*{Concurrent ICL and IWL reproduce both curriculum effects} 
While the above explorations showed how IWL and ICL can produce different curriculum effects, we are now in a position to study how the two might interact in a single system capable of both.
To do this, we took our network that developed ICL abilities through metalearning, and gave it unseen category-learning tasks, allowing it to learn by either ICL (via forward activation dynamics) or IWL (via error backpropagation).
Here, we predicted that the dynamic interaction between IWL and ICL would qualitatively reproduce the full set of curriculum effects observed in the original study \cite{NohYanBjorkEtAl16}: ICL would produce the blocking advantage in the presence of rule-like structure, while IWL would produce the interleaving advantage in the absence of such structure (see principle~\ref{item:both}).

As we described above, when categories are governed by rule-like structure, ICL succeeds on the task and exhibits a blocking advantage in few-shot inference. 
But in the rotated task, where categories are not governed by rule-like structure, ICL struggles (Figure~\ref{fig:category}g). 
The resulting errors drive an increase in IWL, producing an interleaving advantage due to catastrophic forgetting (Figure~\ref{fig:category}i; interaction between curriculum and rotation: $p < 10^{-3}$).

Taken together, the above experiments show that a single model capable of ICL and IWL can recapitulate the curriculum effects observed in human category-learning \cite{NohYanBjorkEtAl16}.
When the network is capable of making inferences over familiar rules, it can solve new tasks from a few samples given in context. 
However, when the environment does not afford such inferences or the network cannot make them, IWL can still compensate, allowing good performance.
This IWL suffers from catastrophic forgetting, resulting in an interleaving advantage on the rotated task.

\subsection*{Curriculum effects in a compositional task}
As noted above, one of the most impressive recent developments in research on neural networks has been the demonstration that ICL can give rise to compositionality \cite{LakeBaroni23a, ZhouScharliHouEtAl22a, RussinMcGrathEtAl2024}, traditionally considered to be a major theoretical challenge to neural networks \cite{FodorPylyshyn88a, Marcus98}.
Recent studies have shown that while standard IWL struggles to reproduce human-like compositional generalization behaviors \cite{LakeBaroni18, KeysersScharliScalesEtAl19, KimLinzen20a}, ICL can appear to compose inferred rules in order to generalize to new inputs \cite{BubeckChandrasekaranEldanEtAl23, LakeBaroni23a, PressZhangMinEtAl23a, ZhouScharliHouEtAl22a}.
Thus, a key goal of our framework is to leverage the distinction between ICL and IWL to provide a unified account of both the compositional generalization behaviors and the curriculum effects observed in humans.
In particular, ICL should account for both the blocking advantage and the compositional generalization behaviors observed in tasks governed by rule-like structure, while IWL accounts for the interleaving advantage observed when such compositional generalization is challenging or impossible.

\begin{figure*}%[tbhp]
\centering
\includegraphics[width=.99\linewidth]{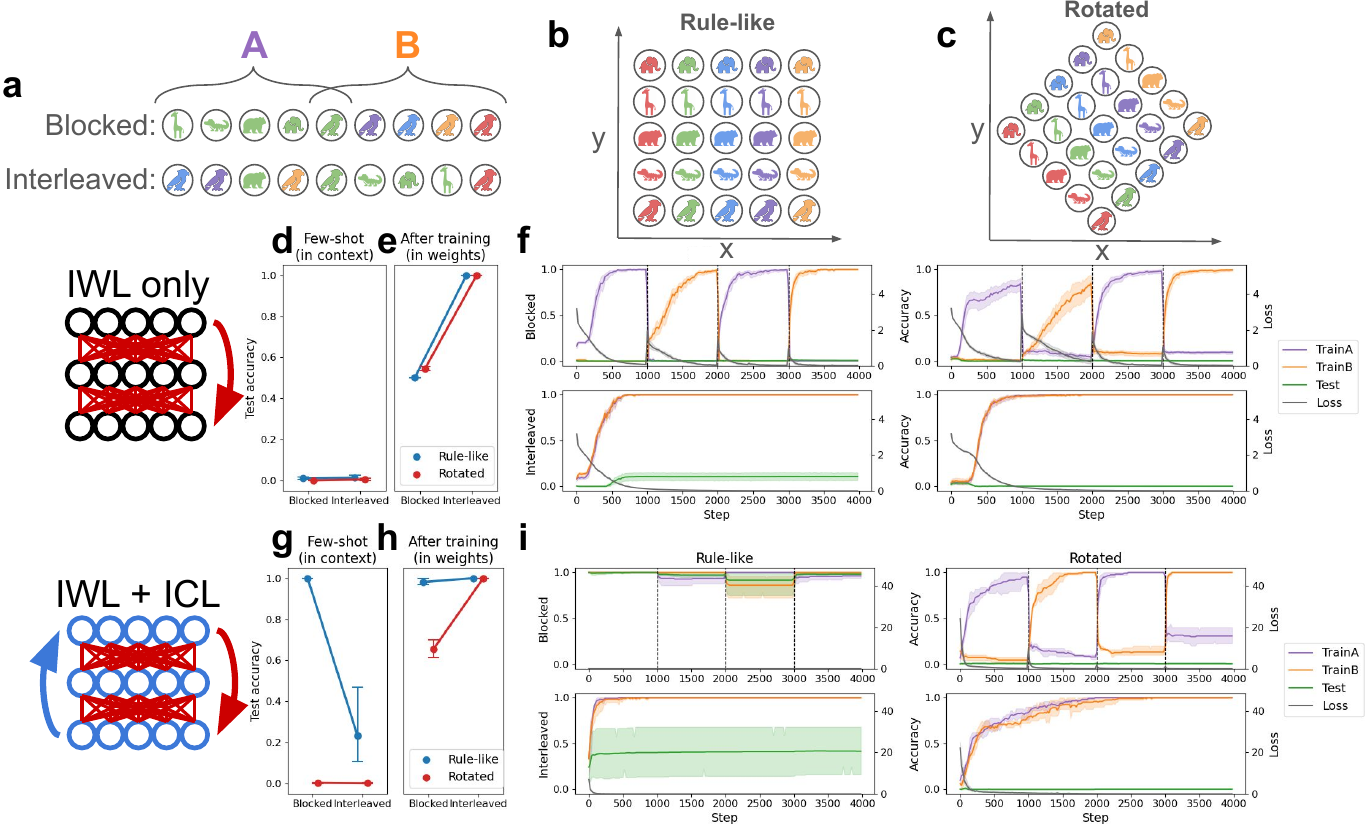}
\caption{Compositional task and results. The task is derived from a human study \cite{DekkerOttoSummerfield22c}. Networks were presented with the locations corresponding to particular cues (colored animals) and had to predict the locations of unseen cues. Cues were comprised of sequences of tokens (e.g., `blue alligator'). See Appendix for details. \textbf{(a)} Curriculum conditions. In both the blocked and interleaved conditions, the 9 study examples always included one full row and one full column. In the blocked condition, the row was presented in full before the column, or vice-versa. In the interleaved condition, these 9 examples were randomly shuffled. \textbf{(b)} In the rule-like condition, locations varied systematically with the color and animal features (e.g., color determined x-coordinate and animal determined y-coordinate). \textbf{(c)} In our novel rotated condition, the original axes were rotated by 45 degrees, so that any change in either color or animal resulted in a change to both x- and y-coordinates. \textbf{(d-f)} Performance on the compositional task with in-weight learning (IWL) only, where again randomly initialized networks were trained from scratch. \textbf{(d)} The few-shot evaluation tested networks' ability to solve the task in context based on the 9 study examples given in the input. Again, without prior metalearning neural networks were incapable of solving the task in this way, regardless of condition. \textbf{(e)} Without prior metalearning experience, the network was still able to learn via IWL, performing well on both the rule-like and the rotated tasks after task-specific training. IWL again exhibited an interleaving advantage due to catastrophic forgetting (see f). \textbf{(f)} Accuracy and loss results over the course of IWL training in each of the four conditions. Accuracy is again split by group, in this case corresponding to whether the cue was part of the row or the column (see a). Here results were similar to the category-learning case, where IWL exhibited catastrophic forgetting when trials were blocked, regardless of rotation condition. IWL also failed to generalize compositionally, failing on the 16 held-out test cues (green lines) in all conditions. \textbf{(g-i)} Experiments using networks capable of both in-weight and in-context learning (ICL). \textbf{(g)} After metalearning, the models again exhibited a blocking advantage, but also showed strong compositional generalization, as shown by the high few-shot test accuracy in the blocked condition. ICL again failed in the rotated condition. \textbf{(h)} After task-specific training, the network exhibited an interleaving advantage in the rotated condition, due to catastrophic forgetting when trials were blocked (see i). \textbf{(i)} When trials were blocked in the rule-like condition, accuracy was near-perfect, resulting in little loss and thus little IWL. In the rotated condition, ICL failed, resulting in larger losses, increased IWL, and increased catastrophic forgetting, as can be seen in the rapid drop in accuracy on the first group (`TrainA,' shown in purple) while training on the second group (`TrainB,' shown in orange), and vice versa. No catastrophic forgetting occurred in the interleaved condition, but compositional generalization (green) was considerably worse than when trials were blocked.}
\label{fig:grid}
\end{figure*}

We focused our investigations on a recent study demonstrating compositional generalization in humans on a novel rule-governed task, where the goal was not to categorize stimuli but to learn a latent compositional coordinate system \cite[see Figure~\ref{fig:grid}a-c;][]{DekkerOttoSummerfield22c}.
Notably, this study showed that compositional generalization indeed depended on the curriculum, improving when related trials were blocked (47/58 participants generalized) compared to interleaved (36/60 participants generalized)---consistent with the idea that the mechanisms underlying compositionality can be linked to those responsible for producing the blocking advantage.
This task therefore provides an excellent testbed for our metalearning neural networks, allowing us to replicate our original curriculum-related results in a different paradigm while also studying their connection to compositionality.

In the original task, participants learned to pair colored animals with arbitrary xy-coordinates via trial-and-error. 
Importantly, the correct locations varied systematically with the two features: color determined the x-coordinate (each of 5 different colors was linked to one of 5 different x-values) while the animal determined the y-coordinate, or vice-versa. 
Participants saw only 9 of the 25 possible color-animal pairs as study examples; they had to make novel inferences to generalize to the 16 remaining pairs during testing (without feedback).
This task can be seen as rule-based in that a simple rule (e.g., color = x, animal = y) governs the locations, and can be seen as compositional in that good test performance requires composition of knowledge about a particular color (e.g., `blue' means x = 3) with knowledge about a particular animal (e.g., `alligator' means y = 2) into a novel combination (e.g., `blue alligator' means location is 3, 2).

The key experimental variable manipulated in the study was the curriculum---which 9 of the 25 cues were used as study examples, and the order in which they were presented (see Figure~\ref{fig:grid}a).
In the \textbf{Blocked} condition, all cues of a particular color (i.e., a single row/column) were presented before all the cues with a particular animal, or vice-versa. 
In the \textbf{Interleaved} condition, a single row and column were again chosen for study, but their order was randomly shuffled.\footnote{Note that the original study also tested two other related conditions, where sampling of items was ``Aligned'' or ``Misaligned''; we simulated  these cases and reproduced similar results in the Appendix, but here focus on the key blocked vs.\ interleaved contrast.}

The experimenters found that human compositional generalization performance depended on which curriculum was used: participants performed better in the blocked than the interleaved condition \cite{DekkerOttoSummerfield22c}.
The original study did not manipulate the presence or absence of rule-like structure as the categorization task did \cite{NohYanBjorkEtAl16}, but we hypothesized that rotating the underlying coordinate grid (see Figure~\ref{fig:grid}c) would cause a similar interleaving advantage to emerge. 
This is because when the underlying coordinate system is rotated, no simple rule (e.g., color = x, animal = y) is available. 
We therefore tested our metalearning models in both the original \textbf{Rule-like} setting, and in a \textbf{Rotated} version.

\subsubsection*{IWL is non-compositional and produces an interleaving advantage} 
As in the simulations with the categorization task, we first evaluated neural networks without ICL capabilities on the task by training them from scratch.
Without ICL, performing the task in the few-shot setting was again impossible (see Figure~\ref{fig:grid}d).
The only way the network could learn was through IWL, which again exhibited an interleaving advantage due to catastrophic forgetting when trials were blocked (confirmed by a main effect of curriculum: $p < 10^{-3}$; see Figure~\ref{fig:grid}e-f).
Because the network had no way of inferring rules from in-context examples, there was no observable difference between the rule-like task and the rotated task.
Furthermore, in both versions of the task the network learned the study examples well when trials were interleaved, but performed poorly on test trials that required compositional generalization. 
Thus, in contrast to the categorization task where the IWL showed good generalization performance (see Figure~\ref{fig:grid}f), the compositional task allowed us to reproduce known failures in compositional generalization in networks capable only of standard IWL \cite{FodorPylyshyn88a, KeysersScharliScalesEtAl19, KimLinzen20a, LakeBaroni18, Marcus98}.

\subsubsection*{ICL can be compositional and can produce a blocking advantage} 
We then endowed the network with ICL abilities by having it metalearn on a distribution of tasks (see methods for details).
After metalearning, these ICL abilities allowed the network to generalize compositionally on unseen tasks, achieving good performance on color-animal combinations that were not included in the study examples.
This generalization performance involved the composition of rules that could be inferred from the study examples (see Figure~\ref{fig:grid}b).
Furthermore, as in the previous simulations, ICL exhibited the same kind of blocking advantage observed in humans \cite{DekkerOttoSummerfield22c}, performing better in the few-shot setting when trials were blocked compared to interleaved (main effect of curriculum on rule-like task: $p < 10^{-3}$).
 
These findings extend recent work \cite{LakeBaroni23a} by showing that the ICL algorithm that emerges in metalearning neural networks can reproduce human-like compositional generalization behavior and its associated blocking advantage in this experimental paradigm \cite{DekkerOttoSummerfield22c}.
This is significant because it shows how neural networks, which have traditionally been criticized for lacking compositionality \cite{FodorPylyshyn88a, Marcus98}, can through metalearning come to implement an ICL algorithm that is capable of human-like compositional generalization \cite{RussinEtAl2024, RussinMcGrathEtAl2024}.

\subsubsection*{ICL and IWL produce compositionality and both curriculum effects}  
Finally, we allowed IWL to occur in the network that was capable of ICL, and replicated the full set of human curriculum effects that we reproduced above in the category-learning setting \cite{NohYanBjorkEtAl16}.
As predicted, ICL failed in our novel rotated version of the task, leading to more errors and thus greater IWL (see Figure~\ref{fig:grid}g).
This increase in IWL led to the emergence of an interleaving advantage (see Figure~\ref{fig:grid}h)---a testable prediction not evaluated in humans in the original study---whereas ICL again produced the blocking advantage in the original rule-like task (see Figure~\ref{fig:grid}g; interaction between rotation and curriculum: $p < 10^{-3}$).
Taken together, our findings on the compositional task are again consistent with our theoretical framework (see principle~\ref{item:both}), and show how the distinction between in-context and in-weight learning can offer a unified account of human compositional generalization capabilities and their dependence on the the learning curriculum \cite{DekkerOttoSummerfield22c}.

\subsection*{LLMs exhibit compositionality and a blocking advantage}
So far, we have established that it is possible for an ICL algorithm to exhibit compositionality and a blocking advantage, and that a single neural network implementing this kind of ICL alongside its usual IWL will reproduce the full set of empirical results that we have been targeting.
A separate question one can ask is \textit{why} a network would develop an ICL algorithm with these particular properties in the first place.
In our metalearning experiments, we used task distributions that promote these properties (see methods), but there may be more naturalistic distributions that could give rise to them.

Although the datasets used for training LLMs are developmentally unrealistic in many ways \cite{Frank23a, Linzen20, WarstadtMuellerChoshenEtAl23}, they are more naturalistic in the sense that they are comprised of natural language text, rather than content that is specifically relevant to our tasks.
These corpora are not purposefully designed to encourage ICL or any of our hypothesized properties to emerge. 
Nevertheless, impressive ICL abilities do arise in these models, giving them the flexibility to accomplish many kinds of tasks in context \cite{BrownMannRyderEtAl20b, BubeckChandrasekaranEldanEtAl23}.
Given the scale and complexity of their training datasets, it is unclear \textit{a priori} what ICL properties LLMs' should develop, but prior work has shown that their emergent ICL abilities can exhibit compositional generalization in some settings \cite{PressZhangMinEtAl23a, WebbHolyoakLu23, ZhouScharliHouEtAl22a}, and can also be sensitive to the order in which in-context examples are provided \cite{ChenChiWangEtAl24, LuBartoloMooreEtAl22}. 

We thus hypothesized that the properties of ICL assumed by our theoretical framework (i.e., compositionality and a blocking advantage, see principle~\ref{item:ICL}) may emerge in LLMs.
We tested this hypothesis by evaluating two LLMs, Llama 2 \cite{TouvronMartinStoneEtAl23} and GPT-3.5 \cite{BrownMannRyderEtAl20b, OuyangWuJiangEtAl22}, on the same compositional task used above. 
We evaluated the emergent ICL abilities of these models by presenting color-animal pairs from the compositional task only in context.\footnote{In principle, these models should also show IWL properties like any other neural network, but it is highly expensive to finetune them, and our main questions here pertain to ICL.}

\begin{figure}%[tbhp]
\centering
\includegraphics[width=.5\linewidth]{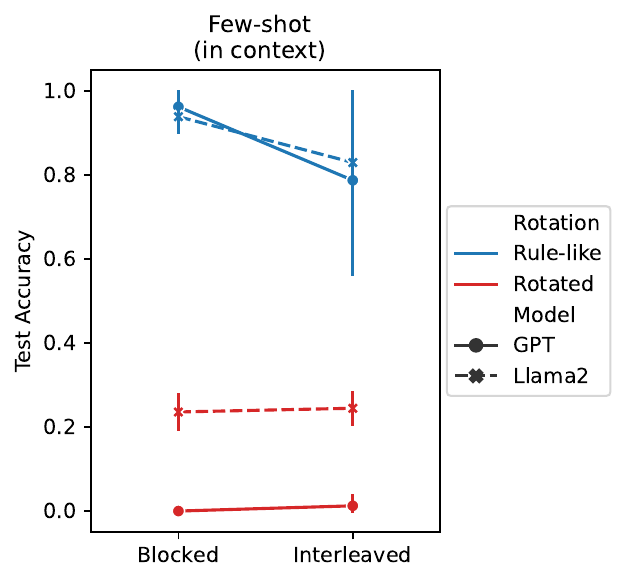}
\caption{LLM results. Large langauge models (LLMs) are capable of in-context learning (ICL) on the text-based version of the compositional task based on the human study \cite{DekkerOttoSummerfield22c}. Both GPT-3.5 (solid lines) and Llama 2 (dashed lines) achieved good compositional generalization performance on the rule-like version of the task (blue), and also exhibited a blocking advantage, performing better when trials were blocked than interleaved (see Figure~\ref{fig:grid}). ICL performance was much worse on the rotated task (red), consistent with our theoretical framework.}
\label{fig:llms}
\end{figure}

Both LLMs showed strong compositional generalization performance on the task (see Figure~\ref{fig:llms}), even though they were only given the 9 study examples and had not been explicitly trained on variants of the task. 
This shows that the emergent ICL abilities in these models can produce the kinds of generalization behaviors that standard IWL in neural networks struggles to achieve (see test accuracy in Figure~\ref{fig:grid}f).

Notably, both LLMs also produced the blocking advantage in the rule-like version of the task (curriculum main effect: $p < 10^{-3}$).\footnote{Like our metalearned neural networks, the LLMs also showed the full pattern of curriculum effects described in the human study, see Appendix for details.} 
This again shows that even though the ICL capability in the LLMs has not been specifically sculpted to produce this blocking advantage, it emerges nonetheless via large-scale next-word (next-token) prediction on large corpora of text.

Finally, both LLMs performed poorly on the rotated task (rotation main effect: $p < 10^{-3}$). 
This is also consistent with our theoretical framework (see principle~\ref{item:both}), which predicts that ICL should be more difficult in the absence of rule-like structure because in-context inferences are more complex. 
IWL would be required to compensate for the failure of ICL on such tasks, as we showed in our metalearning experiments.

Thus, neural networks can come to implement an ICL algorithm with the properties of compositionality, a blocking advantage, and a preference for rule-like structure---even when their training does not specifically target these properties, but consists in next-token prediction on naturalistic text.

\subsection*{Tradeoff between flexibility and retention}

\begin{figure*}[h]
\centering
\includegraphics[width=.99\linewidth]{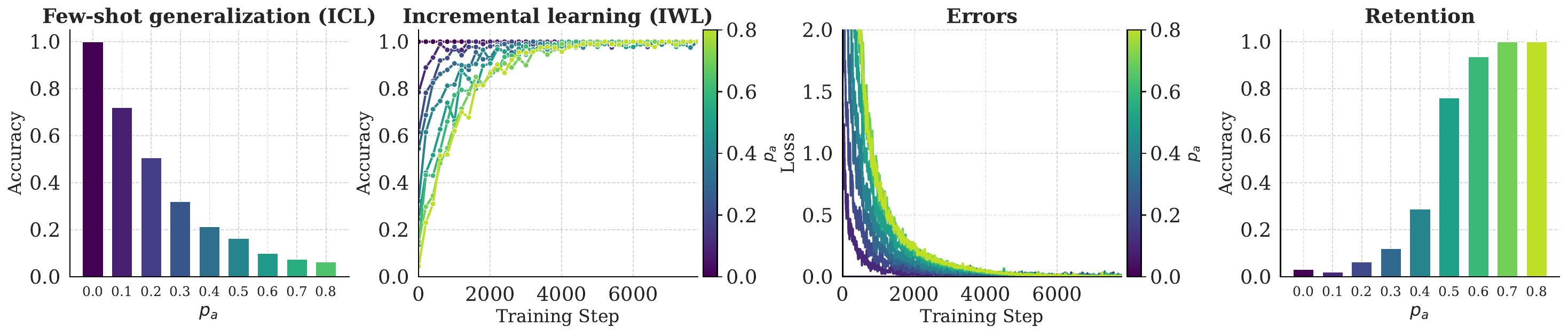}
\caption{Tradeoff between flexibility and retention. Ablating attention to examples in the context (analogous to limiting the number of items accessible in WM) hurts cognitive flexibility but improves retention. When attention is ablated (i.e., $p_a$ is high), few-shot generalization via ICL suffers (left) and incremental learning via IWL is slower (middle left). This results in more errors (middle right), consistent with human EEG data showing prolonged presence of neural signatures of prediction errors under higher WM load \cite{CollinsFrank18b}. When more errors are made, more IWL occurs, resulting in better retention in the absence of contextual information (right), consistent with human results showing better retention after learning under higher WM load \cite{Rac-LubashevskyCremerCollinsEtAl23,HitchcockKimFrank24}. Results are from the compositional task. $p_a$ is the probability that attention to each example in the context was inhibited.}
\label{fig:tradeoff}
\end{figure*}

So far, we have highlighted the advantages of ICL over IWL in supporting rapid few-shot learning and generalization, and only noted that IWL is necessary in cases where ICL is less effective (e.g., when trials are interleaved or in rotated tasks). 
However, an additional benefit of IWL (and relatedly, episodic and semantic memory compared to WM) is that retrieval of information stored in synaptic weights does not require persistent activity (e.g., throughout a delay), and can therefore operate over longer timescales and in the absence of contextual cues \cite{ChanDasguptaKimEtAl22a, Reddy23a}.
Thus, we hypothesized that a natural consequence of the coexistence of ICL and IWL would be the emergence of a tradeoff between \textit{flexibility}, or rapid adaptation to new tasks from only a few examples, and \textit{retention}, or the ability to recall information after longer delays or in the absence of contextual information. 
The key insight is that if learning takes place in a setting where ICL is successful, generalization based on latent rules may be improved, but fewer prediction errors will update weights. 
When latent rules are identified, they should give rise to reduced prediction errors, and this suppression should in turn relate to better generalization but worse retention.

In fact, analogous findings have been reported in two lines of research on reinforcement learning (RL) in humans \cite{CollinsFrank16b, Collins18, CollinsFrank18b, Rac-LubashevskyCremerCollinsEtAl23, HitchcockKimFrank24}.
When the task was structured with a hidden rule that could be inferred from the context, neural reward prediction errors were suppressed, and this suppression predicted better generalization of that rule \cite{CollinsFrank16b}.
In simpler RL tasks, participants also show particularly rapid learning when they can acquire contingencies using an ICL-like WM strategy, achieving near-perfect performance within only a few presentations of each stimulus \cite{CollinsFrank18b}. 
However, when WM load was increased, many more presentations were required to achieve the same performance, consistent with an IWL-like incremental learning strategy relying more on RL.
EEG recordings showed that neural signatures of reward prediction errors were suppressed when WM load was low \cite{CollinsFrank18b} or when the underlying structure of the task had been inferred \cite{CollinsFrank16b}.
Moreover, this neural marker of an ICL-like WM strategy was predictive of both better generalization \cite{CollinsFrank16b} and reduced retention \cite{Rac-LubashevskyCremerCollinsEtAl23}. To date, however, no single model has accounted for both sets of findings. 

We tested whether a similar tradeoff between flexibility and retention would emerge in our neural networks.
As a proxy for WM load, we ablated the networks' ability to maintain contextual information throughout learning by inhibiting attention to each in-context example with probability $p_a$ (similar results were obtained by adding noise to the attention mechanism itself; see Appendix). 
Retention was evaluated by testing networks on examples they had seen during training, but in the absence of any contextual information (analogous to testing after a delay when relevant information is not available in WM).

The results for the compositional task are shown in Figure \ref{fig:tradeoff} (results for the category-learning task can be found in the Appendix). 
In both tasks, we observed a robust tradeoff between flexibility and retention.
When the ablation was stronger ($p_a$ was higher), few-shot generalization via ICL was worse (Figure \ref{fig:tradeoff}, left) and incremental learning via IWL took longer to reach optimal performance (middle left). 
This meant that more errors were made throughout learning (middle right), consistent with the stronger neural signatures of prediction errors observed in humans under higher WM load \cite{CollinsFrank18b, CollinsCiulloFrankEtAl17}.
More errors resulted in greater IWL, leading the model to perform better on the retention test (right), consistent with the superior retention observed in humans that had learned under higher WM load \cite{Rac-LubashevskyCremerCollinsEtAl23,HitchcockKimFrank24}.

Thus, the same principles that allow the networks to reproduce compositional generalizations and curriculum effects can also explain the tradeoff observed in human RL experiments \cite{CollinsFrank18b, Rac-LubashevskyCremerCollinsEtAl23, HitchcockKimFrank24}.
This tradeoff suggests that ICL and IWL have distinct advantages whose relative importance depends on whether flexibility or retention is prioritized.

\section*{Discussion}
Influential theories in cognitive science posit two distinct systems to account for findings suggesting a duality in human learning \cite{AshbyMaddox11, BotvinickRitterWangEtAl19a, DawNivDayan05, Evans08, StBTEvansStanovich13, NohYanBjorkEtAl16, OReillyNairRussinEtAl20, Sable-MeyerBenjaminWatkinsEtAl24, Sloman96a, CollinsFrank18b}. 
Prominent theories leverage distinctions between controlled vs.\ automatic processing \cite{ShiffrinSchneider77, FabioCapriRomano19, OReillyNairRussinEtAl20}, model-based vs.\ model-free reinforcement learning \cite{DawNivDayan05, DawGershmanSeymourEtAl11, SuttonBarto98}, WM in PFC vs.\ striatal synaptic learning \cite{FrankClaus06, FrankBadre12, CollinsFrank18b, Rac-LubashevskyCremerCollinsEtAl23}, system 2 vs.\ system 1 thinking \cite{Kahneman11}, and rule-based vs.\ procedural learning \cite{AshbyMaddox11, NohYanBjorkEtAl16}.
These theories explain why human learning exhibits different phenomena under different conditions.
Here, we have focused on three such phenomena: 1) compositionality 2) curriculum effects, and 3) the tradeoff between flexibility and retention.
Humans are capable of utilizing rule-like structure to generalize compositionally \cite{DekkerOttoSummerfield22c, LakeLinzenBaroni19a, LakeUllmanTenenbaumEtAl17a, FranklinFrank18a, FranklinFrank20a, FranklandGreene20}, and of integrating over multiple dimensions and making arbitrary associations when no rule-like structure is present \cite{AshbyMaddox05a, AshbyMaddox11, NohYanBjorkEtAl16, McClellandMcNaughtonOReilly95}.
In the former case, learning can be rapid and flexible, and tends to benefit when related trials are blocked over time \cite{DekkerOttoSummerfield22c, FleschBalaguerDekkerEtAl18, NohYanBjorkEtAl16}.
In the latter case, it benefits when trials are interleaved \cite{GoodeMagill86, LandinHebertFairweather93, NohYanBjorkEtAl16, RichlandFinleyBjork04}, and can result in improved retention but limited flexibility and generalization.

Our work shows how these phenomena can be explained by a single neural network capable of two qualitatively distinct learning processes.
In particular, we have shown how metalearning can endow a network with a capacity to learn \textit{in context}, and how this capacity can capture compositionality and the blocking advantage on tasks governed by rule-like structure.
ICL operates when tasks are consistent with readily identifiable rules, but can be unsuccessful on tasks lacking such structure,
triggering error-driven IWL and producing an interleaving advantage due to catastrophic forgetting \cite{McCloskeyCohen89, McClellandMcNaughtonOReilly95}.
This dynamic interaction between ICL and IWL naturally recapitulates the tradeoff between flexibility and retention observed in humans: WM can be leveraged to rapidly learn new stimulus-response rules, but causes reductions in neural prediction errors driving incremental reinforcement learning, resulting in worse retention after longer delays  \cite{Rac-LubashevskyCremerCollinsEtAl23, CollinsFrank18b,HitchcockKimFrank24}.

ICL has recently emerged as an important topic of research in machine learning and artificial intelligence \cite{BrownMannRyderEtAl20b, DongLiDaiEtAl24}.
Studies have investigated what kinds of data-distributional properties \cite{ChanSantoroLampinenEtAl22, Reddy23a} or neural architectures \cite{GrazziSiemsSchrodiEtAl24, SushmaTianMesthaEtAl24, ZucchetKobayashiAkramEtAl24} drive its emergence, as well as the kind of learning algorithm it implements \cite{AkyurekSchuurmansAndreasEtAl23, vonOswaldNiklassonRandazzoEtAl23, XieRaghunathanLiangEtAl22}, and the internal circuits underlying it \cite{OlssonElhageNandaEtAl22, HendelGevaGloberson23, ToddLiSharmaEtAl23}.
Here, we link this recent work to cognitive science, showing how the dynamic interplay between ICL and IWL can offer a unified perspective on compositionality and curriculum effects.
Our work complements prominent dual-process theories by showing how two distinct learning processes can coexist (and compete) within a single neural network.

\subsection*{Curriculum Effects}
There has been some debate about whether humans learn better when related content is blocked or interleaved over time, with some studies finding a blocking advantage \cite{BeukersCollinKempnerEtAl24a, DekkerOttoSummerfield22c, FleschBalaguerDekkerEtAl18, NohYanBjorkEtAl16} and others finding an interleaving advantage \cite{GoodeMagill86, LandinHebertFairweather93, NohYanBjorkEtAl16, RichlandFinleyBjork04}. 
There may be multiple factors that distinguish these cases \cite[e.g., between-category and within-category similarity; ][]{CarvalhoGoldstone14b}, but one important variable may be the presence of rule-like structure: humans have been shown to exhibit a blocking advantage when the task is governed by succinct rules, and an interleaving advantage when the task does not afford such rules \cite{DekkerOttoSummerfield22c, NohYanBjorkEtAl16}. 
These effects are explained by a dual-process account in which a rule-based learning system operates by an explicit hypothesis-testing strategy and a procedural learning system operates by incrementally integrating information over time \cite{AshbyMaddox11, NohYanBjorkEtAl16}.
Our work offers a novel perspective on this dual-process account, showing how a similar duality can emerge in neural networks capable of both ICL and IWL.

In our framework, the interleaving advantage arises because of catastrophic forgetting \cite{McCloskeyCohen89}, which is a natural property of IWL in neural networks due to their use of overlapping distributed representations \cite{McClellandMcNaughtonOReilly95}. 
Might this kind of forgetting explain the interleaving advantage observed in humans?
The brain is thought to mitigate catastrophic forgetting through the use of sparse, pattern-separated representations in hippocampus \cite{McClellandMcNaughtonOReilly95, OReillyBhattacharyyaHowardEtAl14a}.
However, this effect is unlikely to be eliminated completely, so a similar principle may still underlie the modest interleaving advantage observed in humans \cite{NohYanBjorkEtAl16}.
Future work could directly investigate the extent to which the interleaving advantage observed in the absence of rule-like structure is due to this kind of forgetting.

The blocking advantage, on the other hand, doesn't emerge by default in standard neural networks, but a number of studies have explored the neural mechanisms that might underlie it. 
For example, a neural network model of rule-based inference and WM in the PFC showed that blocking related trials over time can encourage abstract rule-like representations to emerge in the network's activations \cite{RougierNoelleBraverEtAl05d}. 
More recent work \cite{RussinZolfagharParkEtAl22} showed that a PFC-like neural network augmented with a gating mechanism and a bias for active maintenance produces a blocking advantage on a task involving cognitive maps \cite{ParkMillerNiliEtAl20}.
Related work has shown how a neural network equipped with a specialized Hebbian gating mechanism \cite{FleschNagySaxeEtAl23} can reproduce a blocking advantage observed in humans on an analogous task \cite{FleschBalaguerDekkerEtAl18}.
A similar Hebbian mechanism was then used to explain the blocking advantage observed in the compositional task studied here \cite{DekkerOttoSummerfield22c}.
Another recent study showed how the blocking advantage observed in humans on a next-state prediction task \cite{BeukersCollinKempnerEtAl24a} was reproduced by a neural network model that actively maintained distinct contextual representations over time \cite{GiallanzaCampbellCohen24a}.
Overall, these studies emphasize how a blocking advantage can emerge when inferences are made through forward activation dynamics (i.e., \emph{in context}), such as those made over items maintained in WM in PFC. 

Our theoretical account of the blocking advantage is broadly consistent with previous models of this effect, but has a number of advantages.
First, we have shown how it can emerge in a neural network model that also produces the interleaving adantage on different tasks.
Furthermore, while our framework is consistent with previous models in suggesting that the blocking advantage is related to activation dynamics \cite[e.g., WM in PFC;][]{RougierNoelleBraverEtAl05d, RussinZolfagharParkEtAl22}, we show how these dynamics can be metalearned by training on a distribution of related tasks \cite{WangKurth-NelsonKumaranEtAl18a}, thus providing a conceptual link between these prior models and ongoing work investigating metalearning and cognitive flexibility in natural and artificial intelligence \cite{BinzDasguptaJagadishEtAl23a, LakeBaroni23a, JagadishCoda-FornoThalmannEtAl24, SandbrinkSummerfield24, Wang21, WebbHolyoakLu23}. 

Indeed, we also observed a blocking advantage in LLMs, which have revolutionized artificial intelligence research \cite{BrownMannRyderEtAl20b, BubeckChandrasekaranEldanEtAl23} and appear to exhibit high levels of cognitive flexibility \cite{BubeckChandrasekaranEldanEtAl23, MirchandaniXiaFlorenceEtAl23a}.
These results show that a blocking advantage can emerge with ICL even when networks are trained on natural text rather than metalearning datasets specifically designed to promote it. 
Although it is difficult to know exactly why this blocking advantage emerges in the LLMs, we speculate that it is driven by distributional properties of the natural text corpora on which they are trained, such as the tendency for human writing to afford inferences best made by assimilating consecutive examples in a sequential, rather than haphazard, manner.
However, further work is needed to better understand the sources of the blocking advantage in the LLMs, and the internal mechanisms responsible for producing it.

In general, our work does not directly address whether the blocking advantage observed in humans emerges due to strong constraints imposed by neural architecture (e.g., recurrence, limitations in WM capacity), rather than the statistical properties of the environment (e.g., the distributional properties of natural language).
In our experiments, the metalearning networks and the LLMs utilized the transformer architecture \cite{VaswaniShazeerParmarEtAl17b}, which is not recurrent and does not have hard constraints in WM capacity.
Both the blocking advantage and the preference for rule-like tasks emerged in these models due to the statistical properties of their training data. 
This was especially clear in the metalearning experiments, where we had full control over the data distribution and confirmed that it determines when the blocking advantage emerges (see Appendix). 
Consistent with these findings, prior work has shown that metalearning networks trained on category-learning problems that match the natural statistics of real-world tasks perform poorly on the same problems that humans struggle with \cite{JagadishCoda-FornoThalmannEtAl24}.
Furthermore, the human blocking advantage has been shown to depend on the extent to which the feature dimensions relevant to the rule-like structure of the task are represented in a strongly segregated manner \cite{FleschBalaguerDekkerEtAl18}, a factor that is likely to depend on an individual's prior learning experiences.
However, we think that the human blocking advantage is also likely to depend on key architectural features of the human brain, such as its recurrence and the mechanisms for gating and serial attention in PFC and basal ganglia \cite{OReillyFrank06, FrankBadre12, RougierNoelleBraverEtAl05d, RussinZolfagharParkEtAl22}. 
These, in turn, might affect the distributional properties of natural language that is produced by humans and provided as training data for the LLMs \cite{XuFutrell25}.
Further work is required to understand how architectural features interact with the distributional properties of a network's training data, and how they might impact the emergence of ICL with specific properties.

\subsection*{Compositionality}
Compositionality is thought to be a key property underlying human cognitive flexibility, permitting familiar rules or concepts to be combined in novel ways, thus facilitating a powerful form of generalization \cite{FodorPylyshyn88a, LakeUllmanTenenbaumEtAl17a, RussinMcGrathEtAl2024, OReillyRanganathRussin22}. 
Recent work has shown that although compositionality may not be a natural property of standard IWL in neural networks \cite{FodorPylyshyn88a, LakeBaroni18, KimLinzen20a, Marcus98}, it can emerge as a property of an ICL algorithm \cite{LakeBaroni23a, RussinMcGrathEtAl2024}. 
Our results build on this work, showing that it is possible to endow a neural network with an ICL algorithm that is capable of reproducing the compositional generalization behaviors observed in humans in a recent study \cite{DekkerOttoSummerfield22c}, even when standard IWL fails (see test accuracy in Figure~\ref{fig:grid}f). 
We showed that this kind of ICL algorithm can be metalearned by training on a distribution of related tasks, but also emerges in LLMs trained on large corpora of text (see Figure~\ref{fig:llms}).
While metalearning offers a clear understanding of how a neural network can come to implement an emergent compositional learning algorithm \cite{LakeBaroni23a, RussinMcGrathEtAl2024}, it is less clear why this property would emerge in LLMs trained on next-word prediction.
One suggestion is that at large enough scales, the language modeling objective used in LLMs can itself be seen as engendering a kind of metalearning \cite{BrownMannRyderEtAl20b, SandbrinkSummerfield24}, where some subset of training samples puts pressure on these models to learn how to compose novel concepts or reasoning steps in context \cite{RussinMcGrathEtAl2024}.
This is consistent with the hypothesis that human compositionality is metalearned---a conjecture that, while difficult to study, may yield specific empirical predictions \cite{RussinEtAl2024, LakeBaroni23a, PiantadosiAslin16, PiantadosiPalmeriAslin18}.  
Finally, a key contribution of our work is that it builds on studies linking compositionality to curriculum effects in humans \cite{DekkerOttoSummerfield22c}, providing a unified account of compositional generalization and its dependence on curriculum. 

\subsection*{One network or two systems for category learning?}
While our neural networks are not meant to be comprehensive models of human category learning, they may be relevant to other phenomena observed in category-learning studies.
One ongoing debate in this area has been about whether human category learning is best characterized by a single learning system or by multiple systems \cite{AshbyAlfonso-ReeseTurkenEtAl98, AshbyMaddox05a, AshbyMaddox11, AshbySmithRosedahl20, NosofskyJohansen00, Nosofsky11, MindaRoarkKalraEtAl24, NewellDunnKalish11a, PoldrackFoerde08, StantonNosofsky07}.
Single-system theories emphasize the principle of parsimony, and argue that a system that relies on stimulus similarity and selective attention can explain most of the available findings \cite{NosofskyJohansen00}. 
Multiple-systems theories argue that a single system is not sufficient to account for double dissociations evident in human behavior \cite{AshbyMaddox11, MaddoxFiloteoHejlEtAl04, MilesMatsukiMinda14}, such as the one pertaining to the curriculum effects discussed above \cite{NohYanBjorkEtAl16}.

Our work may help resolve this debate by showing how such double dissociations can be explained by a single network that can learn in two different ways. 
ICL and IWL are not separate learning \textit{systems}, but nevertheless manifest fundamentally different properties and compete to drive learning behavior, with each taking precedence at different times.
Our approach arguably maintains the parsimony of a single-system theory in the sense that these two distinct sets of learning properties emerge from the natural dynamics of a single network, rather than being independently posited as part of separate systems.
However, as discussed above, the properties of ICL and IWL align well with the two systems proposed in prominent multiple-system theories \cite{AshbyAlfonso-ReeseTurkenEtAl98, AshbyMaddox11}, with ICL corresponding to the explicit, verbal system and IWL corresponding to the implicit, procedural system. 

In addition to the curriculum effects we observed in our experiments, the distinction between ICL and IWL may help to explain other findings motivating multiple-system theories of category learning.
For example, some studies have shown that increased WM load can impair rule-based learning \cite[][although see \citealp{NosofskyKruschke02, MindaRoarkKalraEtAl24}]{MaddoxFiloteoHejlEtAl04, MindaRabi15, QuamWangMaddoxEtAl18, WaldronAshby01}. 
This finding parallels our results showing that ICL-mediated generalization suffers when access to contextual information is restricted (see Figure \ref{fig:tradeoff}).
Studies of category learning have also shown that humans are capable of generalizing outside of their training distribution on rule-based tasks, but cannot do so in information-integration (rotated) tasks \cite{CasaleRoederAshby12, GanZhengWangEtAl23}.
This aligns with our finding that ICL, but not IWL, was capable of generalizing on both tasks (see Figure \ref{fig:grid}).
Finally, there is some evidence that children struggle specifically with rule-based category-learning tasks \cite{Huang-PollockMaddoxKaralunas11, RabiMilesMinda15, RabiMinda14, RoarkLeschtHamptonWrayEtAl23}, but can perform at adult levels when categories are based on family resemblances \cite{MindaDesrochesChurch08}.
This is consistent with our neural networks, which are inherently capable of IWL but only develop sophisticated ICL through metalearning \cite{RussinEtAl2024}.

Our models may also clarify certain outstanding questions for current multiple-system theories.
For example, behavioral evidence suggests that the verbal or explicit system operates by default initially in humans, but it is unclear \textit{a priori} why this would be the case \cite{NohYanBjorkEtAl16, AshbyMaddox11}.
In our neural network models, ICL operates by default because it can occur at a much faster timescale (through activation dynamics), and because IWL only occurs when errors are made.
Another unresolved question concerns evidence from neuroimaging studies on category learning suggesting that there is substantial overlap in the brain regions active during rule-based and information-integration tasks \cite{CarpenterWillsBenattayallahEtAl16, MiltonBealingCarpenterEtAl17}. 
This can seem to contradict the predictions of a multiple-system theory that posits completely independent learning modules.
The distinction between ICL and IWL provides a natural explanation for this finding, as these two learning processes coexist throughout the network and therefore need not be localizable to separate regions.

In fact, our neural networks are likely to be unrealistically homogeneous, as they have no inherent modularity at all.
Many findings suggest that specific brain regions such as PFC are particularly important for cognitive functions such as WM, rule-based inference, and modulating processing according to the current context or goal \cite{BuchsbaumGreerChangEtAl05, MillerCohen01, Milner63, RougierNoelleBraverEtAl05d, WallisAndersonMiller01}.
We speculate that the organization of the human PFC, which has an intrinsic bias to robustly maintain information over longer timescales until it is actively updated \cite{OReillyFrank06, HuntHayden17a, CavanaghHuntKennerley20}, may encourage ICL abilities, along with their specific properties, to become partially localized to this area \cite{RussinOReillyBengio20a, RussinZolfagharParkEtAl22}.

Although our models did not contain any separate PFC-like system, we note that the ICL algorithms implemented in their activation dynamics can be seen as analogous to those observed in neural models of PFC trained across multiple tasks \cite{RougierNoelleBraverEtAl05d, WangKurth-NelsonKumaranEtAl18a, CollinsFrank13a}.
Just as in our models, these ICL-like abilities only emerge through IWL-like learning of abstract representations in PFC and of gating policies in BG.
Recent work has shown that transformer architectures can mimic the frontostriatal gating mechanisms in these biological models when trained on human WM tasks, and exhibit effective capacity limitations despite the lack of any inherent architectural constraint imposing such a limitation \cite{TraylorMerulloFrankEtAl24a, SoniTraylorMerulloFrankEtAl24a}.
Future work could use similar techniques to investigate whether emergent PFC-like computational mechanisms also explain ICL-related phenomena in our metalearning networks.

\section*{Methods}

\subsection*{Task details}
 The inputs and outputs of both tasks were encoded into sequences of tokens appropriate for processing by standard transformer architectures \cite{VaswaniShazeerParmarEtAl17b}.
In the category task, each of the two feature dimensions could take any of eight possible values (e.g., `length-1,' `length-2,' ...), and each of the categories was associated with one arbitrary label (`A' or `B'). 
Each of these feature values and category labels was encoded as a separate token. 
Inputs to the model consisted of a set of \textit{study examples} along with a single \textit{test query}, all supplied to the model in context.
32 study examples were given, each of which consisted of an item-label pair, where the item contained two tokens corresponding to the values of the two feature dimensions.
The query came after the study examples, and consisted of a single item without a category label. 

In the compositional task, all of the colors and animals were encoded as separate tokens, as were the x- and y- coordinates.
Again, inputs to the model consisted of a set of study examples and a single query.
In this task, the study examples included 9 item-location pairs, where each item contained a color and an animal and each location contained an x- and a y-coordinate. 

\subsection*{Model details} 
In all models, tokens were embedded using a dictionary of learnable vectors.  
In the metalearning experiments, these embeddings started out as arbitrary random vectors and were optimized by end-to-end backpropagation throughout training.
In the LLMs, real English words were used, allowing the models to leverage semantic knowledge gained through training.
This is similar to the human participants in the original studies, who could leverage existing knowledge that color and animal are orthogonal feature dimensions, for example.

All metalearning experiments used the same transformer architecture \cite{TouvronMartinStoneEtAl23, VaswaniShazeerParmarEtAl17b}.
An informal hyperparameter search was conducted to find a suitable number of layers, hidden size, dropout, and learning rate.
The size of the feedforward layers was always twice the hidden size. 
The best-performing model was selected based on validation accuracy for each task separately. 
In the category task, the best-performing model had 4 layers, 8 heads, a hidden size of 64, and no dropout.
In the compositional task, the best-performing model had 12 layers, 8 heads, a hidden size of 64, and dropout of 0.1. 
Models were evaluated on exact-match accuracy using greedy decoding and teacher forcing. 

In the LLM experiments, we evaluated GPT-3.5 \cite{BrownMannRyderEtAl20b, OuyangWuJiangEtAl22} and Llama 2 \cite{TouvronMartinStoneEtAl23}.
GPT-3.5 is an LLM trained on next-token prediction and finetuned to be more useful in a chat-based interface.
We used the version of Llama 2 that has not been finetuned on instruction data.
In GPT-3.5 (“gpt-3.5-turbo-instruct”), the temperature was set to 0.1, and five runs were performed. 
A maximum of 7 tokens were generated, and no post-processing was done except to strip extra spaces.
Llama 2 is an open-source model with approximately 70 billion parameters.
The model was run using resources from the Center for Computation and Visualization at Brown University.
The model was quantized so it could fit onto 2 gpus.
A number of different prompts for each model were tested, but good performance was achieved with simple prompts containing only the study examples, and the prompts did not qualitatively change the pattern of results across conditions.

\subsection*{Metalearning}
 We adopted a metalearning framework to 
induce ICL abilities to emerge within the activation dynamics of a neural network by training it on a distribution of tasks \cite{WangKurth-NelsonKumaranEtAl18a, BinzDasguptaJagadishEtAl23a, JagadishCoda-FornoThalmannEtAl24, LakeBaroni23a}. 
These task distributions encouraged the resulting ICL algorithm to have a preference for related trials to be blocked over time, and a tendency to generalize compositionally.

For the category-learning experiments, we trained our networks on a distribution of tasks with the same basic structure described above. 
Each individual task was sampled as follows:
2 feature dimensions were sampled uniformly without replacement from a set of 200 unique dimensions. 
Each of these dimensions had 8 possible values, making 64 possible items in the newly sampled task.
One of two possible category labels was randomly assigned to each of the two categories.
In each new task, 16 items from each category were randomly chosen to be included in the set of 32 study examples. 
The queries seen during metalearning could either be one of the 32 given in the context (``train''), or one of the remaining 32 (``test'').  
In our main experiments, all samples in the metalearning distribution used the rule-like task and the blocked condition.
The network metalearned on 12,000 tasks sampled in this way, and was subsequently tested on a held-out set of 100 tasks with combinations of dimensions that had not been seen during training. 
A further 10 held-out tasks were used for testing.
During metalearning in the category setting, networks trained for 20 epochs with cross-entropy loss, the Adam optimizer \cite{KingmaBa15}, a learning rate of 0.0001, and a batch size of 256.

We also constructed a metalearning distribution based on the design of the compositional task \cite{DekkerOttoSummerfield22c}. 
Again, each individual task in this distribution had the same structure as the compositional task presented above.
The tasks were sampled as follows:
First, the orders of the lists of five colors and five animals were shuffled, determining their corresponding orders in the 5x5 grid of locations. 
Then, the two features were randomly assigned to the x- and the y-coordinates (color = x and animal = y, or vice versa). 
In the rotated condition, this 5x5 grid was rotated by 45 degrees and scaled so that each coordinate of each cue landed on an integer. 
As in the category-learning setting, all samples in the metalearning distribution were rule-like and blocked.
We again generated 12,000 tasks for metalearning, and used 100 held-out tasks with different 5x5 grids for validation. 
A further 10 held-out tasks were used for testing.
During metalearning in the compositional task setting, networks trained for 500 epochs with the Adam optimizer \cite{KingmaBa15}, a learning rate of 0.001, and a batch size of 256.

\subsection*{Task-specific training}
Once the network acquired an ICL algorithm through metalearning, it was subsequently evaluated on its ability to learn new unseen tasks from each condition. 
This evaluation was conducted in two ways. 
In the \textbf{few-shot} evaluation, the weights of the network were frozen, ensuring that all learning was due to ICL on the study examples given in context. 
In \textbf{task-specific training}, the model's weights were not frozen, and any errors made were used to update weights. 
During task-specific training, the model learned a single task and only received feedback on the study examples, thus emulating the experience of the human participants \cite{DekkerOttoSummerfield22c}.  
Note that this is unlike the metalearning phase, when the model learned how to generalize to queries not included in the study examples.
This second task-specific learning phase that the model underwent can be understood as `finetuning' the model on a specific task, while the metalearning can be understood as `pretraining.'
During task-specific training, networks were again trained with cross-entropy loss and the Adam optimizer \cite{KingmaBa15}, with a learning rate of 0.00001 in the category-learning task, and a learning rate of 0.0001 in the compositional task.
In both tasks, the batch size was equal to the total number of examples (i.e., queries) used in a given block (32 in the category-learning setting, 5 in the compositional setting).

During the task-specific training phase, samples were either blocked or interleaved in two distinct but congruent ways. 
In the blocked condition, related items were blocked over the context, but they were also blocked over the gradient steps (i.e., the model was trained for N gradient steps on samples containing queries from one stimulus group, then was trained for N gradient steps on samples containing queries from the other group, and so on). 
Likewise, in the interleaving condition, items from each group were interleaved both over the context and over the gradient steps.
In the main experiments, the curriculum condition was always consistent during task-specific training---related items were either blocked over both the context and the gradients steps, or interleaved over both the context and the gradient steps. 
However, for the sake of completeness we experimented with all combinations and report these results in the Appendix.

\subsection*{Acknowledgments} 
We would like to thank all members of the Language Understanding and Representation Lab and the Laboratory of Neural Computation and Cognition at Brown University, as well as the Analogy and Attentional Control groups for helpful discussions. MJF was supported by ONR grant N00014-23-1-2792. EP and JR were supported by NIH NIGMS COBRE grant \#5P20GM103645-10.

\newpage
% \section*{Appendix}
\appendix
\section{Extended Methods}

\subsection{Aligned and misaligned curricula in compositional task}
\begin{figure}[h]
\centering
\includegraphics[width=\textwidth]{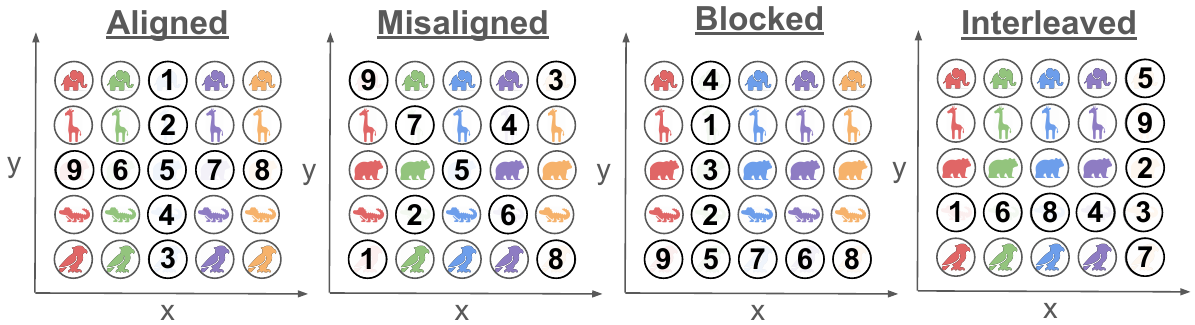}
\caption{Aligned and misaligned curriculum conditions. In the original experiment \cite{DekkerOttoSummerfield22c}, the compositional task also included two additional curriculum conditions. In the aligned condition, participants saw the entire middle row of the grid before seeing the entire middle column (or vice versa). This condition is the same as the blocked condition, except that the row and column used for the study examples were constrained to be the middle row/column. In the misaligned condition, participants saw the cues from one of the diagonals of the grid, followed by cues from the other diagonal.}
\label{fig:aligned_task}
\end{figure}

In addition to the blocked and interleaved curricula presented in the main text, the experimenters who designed the original task \cite{DekkerOttoSummerfield22c} also tested humans on two additional curriculum conditions (see Figure \ref{fig:aligned_task}). 
In the aligned condition, cues from the middle row of the grid were presented first, followed by cues from the middle column (or vice versa). 
This condition was equivalent to the blocked condition, except that the row and column of the grid used for training were always the middle row and middle column.
In the misaligned condition, cues from each of the two diagonals of the grid were presented one after another.
All experiments reported in the main text only used the blocked and interleaved conditions, but we also assessed our metalearning networks and the large language models (LLMs) on the aligned and misaligned conditions (see Additional Results below). 

\subsection{Text-based versions of the tasks}

\begin{figure}
\centering
\includegraphics[width=\textwidth]{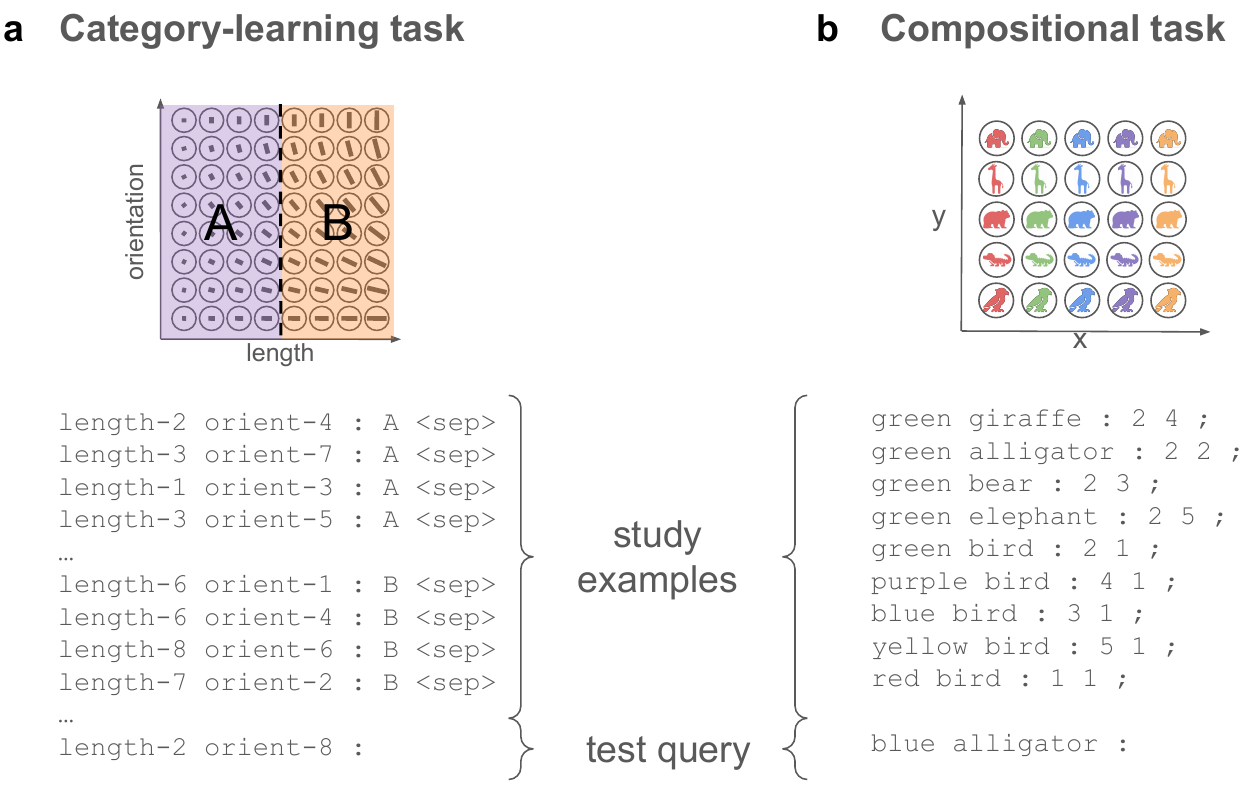}
\caption{Text-based version of the category-learning task (a) and the compositional task (b). In the metalearning experiments, arbitrary strings were used for both tasks (not shown for the compositional task), while in the LLM experiments on the compositional task, common English colors and animals were used. Note that study examples and query appear on new lines in the figure for illustration purposes, but in the experiments examples were only separated by `<sep>' or by semicolons.}
\label{fig:text_tasks}
\end{figure}

Text-based versions of both the category-learning task \cite{NohYanBjorkEtAl16} and the compositional task \cite{DekkerOttoSummerfield22c} were developed for use with standard transformer architectures \cite{VaswaniShazeerParmarEtAl17b}.
A space tokenizer was used in all metalearning experiments.
In the category-learning task (see Figure \ref{fig:text_tasks}a), each of the two dimensions of the stimuli (e.g., line orientation and line length) were coded with a separate set of 8 tokens (e.g., `length-1', `length-2', ...; `orientation-1', `orientation-2', ...), and each of the 2 category labels was coded with a separate token (`A' and `B'). 
Each of the 32 examples given to the model in context contained a stimulus item (i.e., a line with a particular length and orientation), paired with its associated category label via a colon.
Examples were separated by a special `<sep>' token (e.g., `length-6 orientation-3 : A <sep> length-7 orientation-2 : A ...').
The query was appended to this context full of 32 study examples, and consisted of a single item and a colon.  

In the compositional task (see Figure \ref{fig:text_tasks}b), again each of the two dimensions of the stimuli (i.e., the color and animal) were coded with a separate set of 5 tokens.
In the metalearning experiments, arbitrary strings (e.g., `color-1', `color-2') were used, but in the experiments with LLMs, we used a set of common English words for the colors (e.g., `red', `green', etc.) and the animals (`bear`, `elephant', etc.). 
The x and y coordinates of the locations corresponding to each cue were coded with natural numbers (e.g., `1', `2', ...). 
Each of the 9 study examples contained a cue paired with its location; examples were again separated by a special token in the metalearning experiments (`<sep>'), but were separated by a semicolon in the LLM experiments. 

\subsection{Training details}
Metalearning networks were trained on distributions of tasks and evaluated on tasks that were not seen during training.
In both the metalearning and the task-specific training phases, networks received text-based inputs like the ones described above and predicted the outputs corresponding to the test query (i.e., the 1-token category label following the test query in the category-learning task, or the 2-token reward location following the test query in the compositional task). 
Cross-entropy loss was computed at only these specific output tokens.
During metalearning, models were trained with the Adam optimizer

In both task settings, 10 different runs of metalearning were performed with different random initializations.
Unless otherwise specified (see Additional Experiments below), metalearning was always done with task distributions that had rule-like structure and blocked curricula. 
After metalearning was complete, models were evaluated in the few-shot setting on 100 validation tasks and independently underwent task-specific training (finetuning) on 10 different test tasks from each of the four conditions (Rule-like and Blocked, Rule-like and Interleaved, Rotated and Blocked, Rotated and Interleaved).
During task-specific training,  
Particular seeds were excluded if they did not pass a 90\% accuracy threshold in the few-shot evaluation, because our main questions of interest pertained to models that had acquired a competence for in-context learning. 
In the IWL-only experiments (see Figure 2d-f and Figure 3d-f in the main text), the metalearning step was skipped, and 10 different randomly initialized networks underwent task-specific training on 10 separate test tasks from each condition in the same way.

\subsection{Tradeoff between flexibility and retention}
In the experiments studying the emergent tradeoff between flexibility and retention (Figure 5 in the main text), we restricted the amount of contextual information available to the model by ``ablating'' attention to the examples available in the context.
This was achieved by masking attention to each of these examples independently with probability $p_a$.
Masking tokens in this way sets the attention distributions over them to 0, thus disallowing representations encoding subsequent positions in the sequence from extracting any information from them.
In our experimental setup, this was analogous to increasing working memory load or reducing working memory capacity, because when attention to particular examples is masked, they can no longer be used to inform the models' predictions for the query.
We also tried applying noise to the attention mechanism (the value vectors) itself, which produced qualitatively similar results (see Additional results section below). 
Attention was ablated at different levels of $p_a$ throughout the task-specific training of models that had already undergone metalearning on unrotated tasks with interleaved curricula.
In the flexibility test (leftmost plot in Figure 5 of main text), accuracy on held-out examples was tested under those same values for $p_a$.
In the retention test (rightmost plot in Figure 5 of the main text), models were evaluated on seen examples without any contextual information available ($p_a = 1.0$). 
This simulated a test after a long delay, where no information stored in working memory could be leveraged to make predictions.
We tested this tradeoff in both the category-learning task and the compositional task, but due to space limitations, we only report the results for the compositional task in the main text. 
The results for the category-learning task are shown below (see Additional results section).

\section{Statistical testing}

\subsection{IWL exhibits an interleaving advantage on the category task}
We hypothesized that simple in-weight learning on the category-learning task would show an interleaving advantage, regardless of rotation. 
To test this hypothesis, we trained networks from scratch in each of the four conditions (Rule-like and Blocked, Rule-like and Interleaved, Rotated and Blocked, Rotated and Interleaved) using the same setup used for task-specific training following metalearning.
We analyzed the results using a generalized linear model (GLM) with a quasi-binomial family using the logit link function (no random effects). 
This model predicted the train accuracy (successes vs. failures) across the four experimental conditions from the Rotation condition (Rule-like vs.\ Rotated), the Curriculum condition (Blocked vs.\ Interleaved), and their interaction (see Figure 2e in the main text). 
The results revealed a significant main effect of Curriculum ($ \chi^2 = 786.88$, $p < 2.2e-16$) and a significant interaction between Rotation and Curriculum ($\chi^2 = 31.18$, $p = 2.349e-08$). 
The main effect of Rotation was not significant ($\chi^2 = 0.0$, $p = 1.0$).
The significant interaction was due to a small but statistically significant difference in the final interleaved accuracy between the two rotation conditions, indicating that 100\% accuracy was slightly harder to achieve on the rotated version of the task (see Table \ref{tab:iwl_category}).
We therefore tested the simple main effects of Curriculum in each of the Rotation conditions, and found a significant interleaving advantage in both the Rule-like ($\chi^2 = 7e+16$, $p < 2.2e-16$) and Rotated ($\chi^2 = 393$, $p < 2.2e-16$) conditions. 
Overall, these findings confirm our hypothesis that in the absence of in-context learning, in-weight learning shows an interleaving advantage that is largely robust to the rotation condition. 

\begin{table}[h]\centering
\caption{IWL interleaving advantage on category-learning task.}
\begin{tabular}{ccrrr}
Rotation & Curriculum & Successes & Failures & Accuracy \\
\midrule
Rule-like & Blocked & 1600 & 1600 & 50.0\% \\
Rule-like & Interleaved & 3200 & 0 & 100.0\% \\
Rotated & Blocked & 1600 & 1600 & 50.0\% \\
Rotated & Interleaved & 3134 & 66 & 97.9\% \\
\bottomrule
\end{tabular}
\label{tab:iwl_category}
\end{table}

\subsection{ICL exhibits a blocking advantage on the category task}
To test the hypothesis that ICL would show a blocking advantage on the rule-like version of the category-learning task, we evaluated the few-shot test accuracy of metalearned networks on all four conditions (see Table \ref{tab:icl_category} and Figure 2g in the main text). 
We again analyzed results using a GLM with a quasi-binomial family and logit link function.
This test revealed a significant interaction between Rotation and Curriculum ($\chi^2 = 4.91$, $p= 0.027$) and a significant main effect of Rotation ($\chi^2 = 29.86$, $p= 4.65e-08$), but no significant main effect of Curriculum ($\chi^2 = 1.16$, $p= 0.28$).
Due to the presence of a significant interaction, we tested the simple main effect of Curriculum in the Rule-like condition, which showed a significant blocking advantage ($\chi^2 = 5.51$, $p= 0.019$), confirming our hypothesis.

\begin{table}[h]\centering
\caption{ICL blocking advantage on category-learning task.}
\begin{tabular}{ccrrr}
Rotation & Curriculum & Successes & Failures & Accuracy \\
\midrule
Rule-like & Blocked & 3172 & 28 & 99.1\% \\
Rule-like & Interleaved & 2844 & 356 & 88.9\% \\
Rotated & Blocked & 2269 & 931 & 70.9\% \\
Rotated & Interleaved & 2009 & 1191 & 62.8\% \\
\bottomrule
\end{tabular}
\label{tab:icl_category}
\end{table}

\subsection{Concurrent ICL and IWL reproduce both curriculum effects on the category task}
Our main hypothesis was that a single network capable of both ICL and IWL would reproduce the interaction between Rotation and Curriculum observed in humans in the category-learning setting \cite{NohYanBjorkEtAl16}, with ICL producing a blocking advantage in the Rule-like task and IWL producing an interleaving advantage in the Rotated task. 
We evaluated metalearned networks' few-shot test accuracy on the Rule-like task (see Figure 2g in the main text) and their train accuracy after task-specific training on the Rotated task (see Figure 2h in the main text), and performed the same GLM analysis (see Table \ref{tab:icl_iwl_category}).
We observed significant main effects of both Rotation ($\chi^2 = 52.23$, $p= 2.96e-13$) and Curriculum ($\chi^2 = 60.58$, $p= 7.07e-15$), as well as a significant interaction ($\chi^2 = 52.19$, $p= 5.03e-13$). 
Analyses of the simple main effects revealed a significant blocking advantage in the Rule-like task (same as ICL result above; $\chi^2 = 5.51$, $p= 0.019$), and a significant interleaving advantage in the Rotated task ($\chi^2 = 911$, $p < 2.2e-16$).
These analyses confirmed our hypothesis that it is possible for a single neural network model to reproduce the blocking advantage observed in humans on the Rule-like task (due to ICL) and the interleaving advantage observed in humans on the Rotated task (due to IWL). 

\begin{table}[h]\centering
\caption{Concurrent ICL and IWL on category-learning task.}
\begin{tabular}{ccrrr}
Rotation & Curriculum & Successes & Failures & Accuracy \\
\midrule
Rule-like & Blocked & 3172 & 28 & 99.1\% \\
Rule-like & Interleaved & 2844 & 356 & 88.9\% \\
Rotated & Blocked & 1976 & 1224 & 61.8\% \\
Rotated & Interleaved & 3200 & 0 & 100.0\% \\
\bottomrule
\end{tabular}
\label{tab:icl_iwl_category}
\end{table}

\subsection{IWL exhibits an interleaving advantage on the compositional task}
The same analyses described above were performed for models in the compositional setting \cite{DekkerOttoSummerfield22c}, where we hypothesized that the same curriculum effects would emerge for ICL and IWL on the Rule-like and Rotated versions of the task (see Figure 3 in the main text).
To test our hypothesis that IWL would by default exhibit an interleaving advantage, we again trained networks from scratch on the task and evaluated their training accuracy in each of the four conditions (see Figure 3e in the main text). 
Again a GLM (quasi-binomial family, logit link function) revealed a main effect of Curriculum ($\chi^2 = 11510.0$, $p < 2e-16$), where IWL exhibited an interleaving advantage regardless of Rotation (see Table \ref{tab:iwl_compositional}).
A main effect of Rotation was also observed ($\chi^2 = 103.9$, $p < 2e-16$), but there was no interaction ($\chi^2 = 0.0$, $p = 1$).

\begin{table}[h]\centering
\caption{IWL interleaving advantage on compositional task.}
\begin{tabular}{ccrrr}
Rotation & Curriculum & Successes & Failures & Accuracy \\
\midrule
Rule-like & Blocked & 361 & 449 & 44.6\% \\
Rule-like & Interleaved & 810 & 0 & 100.0\% \\
Rotated & Blocked & 411 & 399 & 50.7\% \\
Rotated & Interleaved & 810 & 0 & 100.0\% \\
\bottomrule
\end{tabular}
\label{tab:iwl_compositional}
\end{table}

\subsection{ICL exhibits a blocking advantage on the compositional task}
The same GLM analyses were performed to assess whether ICL performed better on the Rule-like version of the compositional task, and whether it exhibited a blocking advantage in the Rule-like task (see Figure 3g in the main text and Table \ref{tab:icl_compositional} below). 
A model predicting performance on the task from Rotation, Curriculum, and their interaction showed a main effect of Rotation ($\chi^2 = 134.7$, $p < 2e-16$), but no main effect of Curriculum ($\chi^2 = 0.035$, $p = 0.85$), and no interaction ($\chi^2 = 1.33$, $p =0.25$).
A follow-up analysis revealed a simple main effect of Curriculum in the Rule-like condition ($\chi^2 = 40.6$, $p = 1.9e-10$), with ICL performing better when trials were blocked compared to interleaved.
These analyses confirmed our hypothesis that ICL would demonstrate better performance and a blocking advantage in the Rule-like condition. 

\begin{table}[h]\centering
\caption{ICL blocking advantage on compositional task.}
\begin{tabular}{ccrrr}
Rotation & Curriculum & Successes & Failures & Accuracy \\
\midrule
Rule-like & Blocked & 480 & 0 & 100.0\% \\
Rule-like & Interleaved & 112 & 368 & 23.3\% \\
Rotated & Blocked & 2 & 478 & 0.42\% \\
Rotated & Interleaved & 1 & 479 & 0.21\% \\
\bottomrule
\end{tabular}
\label{tab:icl_compositional}
\end{table}

\subsection{Concurrent ICL and IWL reproduce both curriculum effects on the compositional task}
Again our main hypothesis about curriculum effects in the compositional task was that a network capable of both ICL and IWL would exhibit an interaction between Rotation and Curriculum, where ICL would produce a blocking advantage on the Rule-like task and IWL would produce an interleaving advantage on the Rotated task (see Figure 3g-i in the main text and Table \ref{tab:icl_iwl_compositional} below). 
This hypothesis was confirmed with an analysis equivalent to the one performed for the category-learning task: the same GLM showed a significant main effect of Rotation ($\chi^2 = 24.7$, $p = 6.6e-07$), a significant main effect of Curriculum ($\chi^2 = 17.1$, $p = 3.5e-05$), and a significant interaction ($\chi^2 = 71.4$, $p < 2.2e-16$).
Follow-up analyses revealed a significant simple main effect of Curriculum in the Rule-like task ($\chi^2 = 40.61$, $p = 1.9e-10$), where ICL showed a blocking advantage, and a significant simple main effect of Curriculum in the Rotated task ($\chi^2 = 726.58$, $p < 2.2e-16$), where IWL showed an interleaving advantage.
Overall, these statistical analyses confirmed our hypothesis that a neural network capable of both ICL and IWL would reproduce human-like curriculum effects on the compositional task \cite{DekkerOttoSummerfield22c}.

\begin{table}[h]\centering
\caption{Concurrent ICL and IWL on compositional task.}
\begin{tabular}{ccrrr}
Rotation & Curriculum & Successes & Failures & Accuracy \\
\midrule
Rule-like & Blocked & 480 & 0 & 100.0\% \\
Rule-like & Interleaved & 112 & 368 & 23.3\% \\
Rotated & Blocked & 170 & 100 & 63.0\% \\
Rotated & Interleaved & 270 & 0 & 100.0\% \\
\bottomrule
\end{tabular}
\label{tab:icl_iwl_compositional}
\end{table}

\subsection{ICL in LLMs performs better on rule-like task and shows blocking advantage}
An alternative method for inducing ICL in neural network models is to train them to predict the next token on large corpora of text \cite{BrownMannRyderEtAl20b, BubeckChandrasekaranEldanEtAl23}.
We evaluated the ICL abilities of two LLMs trained in this way -- Llama 2 \cite{TouvronMartinStoneEtAl23} and GPT-3.5 \cite{BrownMannRyderEtAl20b, OuyangWuJiangEtAl22} -- by testing them on the compositional task (see Figure 4 in the main text).
Statistical tests analogous to those performed for the metalearning experiments were conducted to evaluate our hypotheses that LLMs would demonstrate better ICL performance when the task was rule-like, and when related trials were blocked rather than interleaved (see Table \ref{tab:llms} below).
A GLM (quasi-binomial family, logit link function) revealed a significant main effect of Rotation ($\chi^2 = 428.6$, $p < 2.2e-16$), a significant main effect of Curriculum ($\chi^2 = 85.63$, $p < 2.2e-16$), and a significant interaction ($\chi^2 = 62.2$, $p = 2.0e-13$). 
Further testing was done for each model individually.
Llama 2 showed a main effect of Rotation ($\chi^2 = 377.5$, $p < 2.2e-16$), a main effect of Curriculum ($\chi^2 = 86.2$, $p < 2.2e-16$), and a significant interaction ($\chi^2 = 60.8$, $p = 4.0e-13$).
In the rule-like condition, Llama 2 showed a significant simple main effect of Curriculum ($\chi^2 = 10.7$, $p = 0.001$), performing better when related trials were blocked compared to interleaved. 
We collected less data on GPT-3.5, but the results of statistical testing were qualitatively similar. 
GPT-3.5 showed a significant main effect of Rotation ($\chi^2 = 115.0$, $p < 2.2e-16$), but no significant main effect of Curriculum ($\chi^2 = 1.74$, $p = 0.63$) and no significant interaction ($\chi^2 = 3.0$, $p = 0.39$).
Follow-up analyses revealed a marginal but not statistically significant simple main effect of Curriculum in the rule-like condition ($\chi^2 = 2.6$, $p = 0.1$).
Overall, our LLM experiments showed that, as expected, ICL in these models performs better on tasks governed by rule-like structure and when related trials are blocked over time. 

\begin{table}[h]\centering
\caption{LLM results on compositional task.}
\begin{tabular}{cccrrr}
Model & Rotation & Curriculum & Successes & Failures & Accuracy \\
\midrule
Llama 2 & Rule-like & Blocked & 601 & 39 & 93.91\% \\
Llama 2 & Rule-like & Interleaved & 478 & 98 & 82.99\% \\
Llama 2 & Rotated & Blocked & 151 & 489 & 23.59\% \\
Llama 2 & Rotated & Interleaved & 141 & 435 & 24.48\% \\
\midrule
GPT-3.5 & Rule-like & Blocked & 77 & 3 & 96.25\% \\
GPT-3.5 & Rule-like & Interleaved & 63 & 17 & 78.75\% \\
GPT-3.5 & Rotated & Blocked & 0 & 80 & 0.00\% \\
GPT-3.5 & Rotated & Interleaved & 1 & 79 & 1.25\% \\
\bottomrule
\end{tabular}
\label{tab:llms}
\end{table}

\section{Additional results}

\subsection{Aligned and misaligned curricula in compositional task}
As noted above, we also tested our metalearned models on the aligned and misaligned curriculum conditions (see Figure \ref{fig:aligned_task}) from the original experiment \cite{DekkerOttoSummerfield22c}.
These conditions manipulate not only the order in which examples are presented, but also which particular examples are used for training.
In both of these conditions, the rules governing the task remain the same (e.g., color determines the x-coordinate and animal determines the y-coordinate), but it may be more difficult to extract these rules from the study examples given in the misaligned condition.
For example, if given only study examples from one of the two diagonals, it would be impossible to infer whether color determines the x-coordinate and animal the y-coordinate or vice versa. 
This is because the cues on the diagonal of the latent grid each varies in both the color and animal dimensions (and therefore in both the x and y coordinates). 
In the aligned condition, on the other hand, one of the two features is held constant within a block, allowing the subject to observe how the locations change as a single feature is changed.

\begin{figure}[h]
\centering
\includegraphics[width=0.8\textwidth]{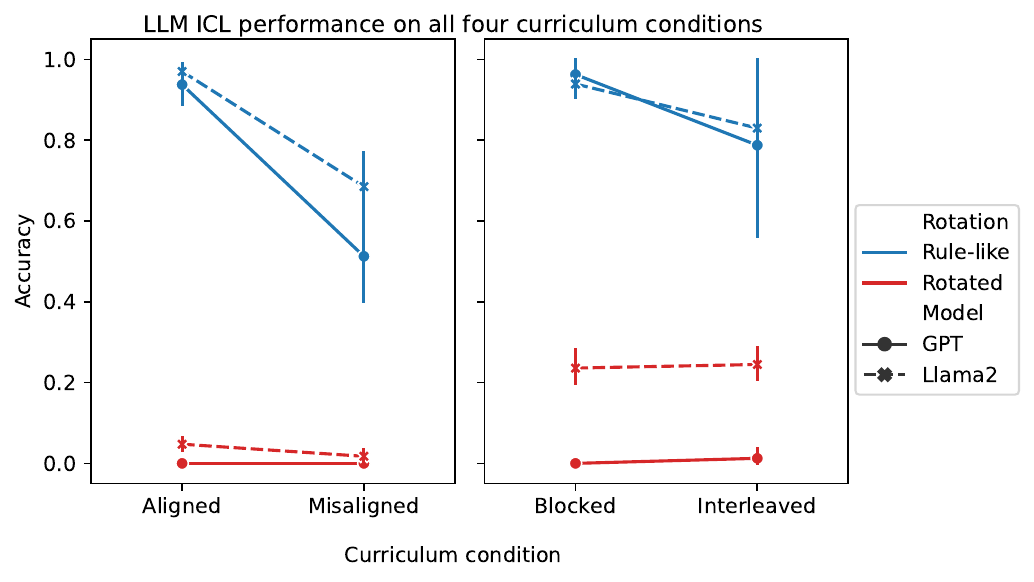}
\caption{ICL performance by LLMs on all four curriculum conditions used in the original human experiments \cite{DekkerOttoSummerfield22c}. Consistent with the human data, both LLMs generalize better in the rule-like task (blue) in the aligned condition than in the misaligned condition, and better in the blocked condition than in the interleaved condition. ICL performance was poor in the rotated task (red), regardless of curriculum condition.}
\label{fig:llm_aligned}
\end{figure}

\begin{figure}[h]
\centering
\includegraphics[width=0.6\textwidth]{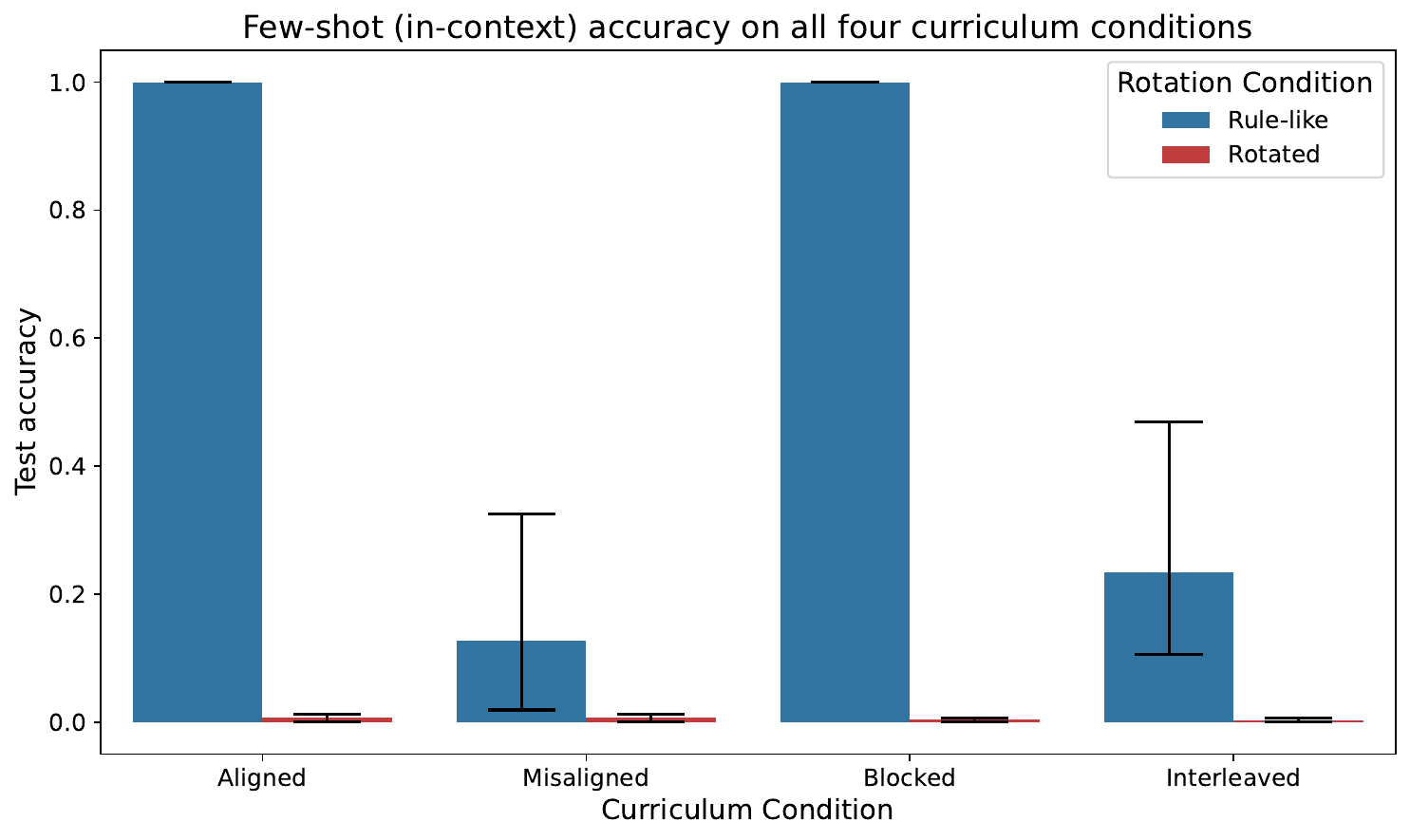}
\caption{In-context learning results for metalearned networks on all four curriculum conditions used in the original human experiments \cite{DekkerOttoSummerfield22c}. In the main text we report results for the blocked and interleaved conditions due to space considerations and because that was our main contrast of interest. However, the same metalearning networks trained on the same distribution of tasks also produce the effects observed in human participants in the aligned and misaligned conditions, generalizing better in the aligned compared to the misaligned condition. The networks perform very poorly on the rotated version of the task, regardless of curriculum (not tested in the original human study), consistent with our hypotheses.}
\label{fig:aligned_results}
\end{figure}

In the human experiments \cite{DekkerOttoSummerfield22c}, participants showed greater compositional generalization on the aligned condition compared to the misaligned condition.
We therefore hypothesized that ICL would show the same effect in both our metalearned networks and in the LLMs.
We tested networks that were metalearned in the same setup described for our main experiments on the aligned and misaligned conditions.
Consistent with our hypothesis, ICL in these networks exhibited better compositional generalization performance in the aligned condition than in the misaligned condition (see Figure \ref{fig:aligned_results}).
The LLMs also showed better generalization in the aligned condition compared to the misaligned condition (see Figure \ref{fig:llm_aligned}).
Taken together, these results suggest that humans and neural networks capable of ICL succeed at generalizing compositionally in similar curriculum conditions.

\subsection{Metalearning on rotated task}
\begin{figure}[h]
\centering
\includegraphics[width=\textwidth]{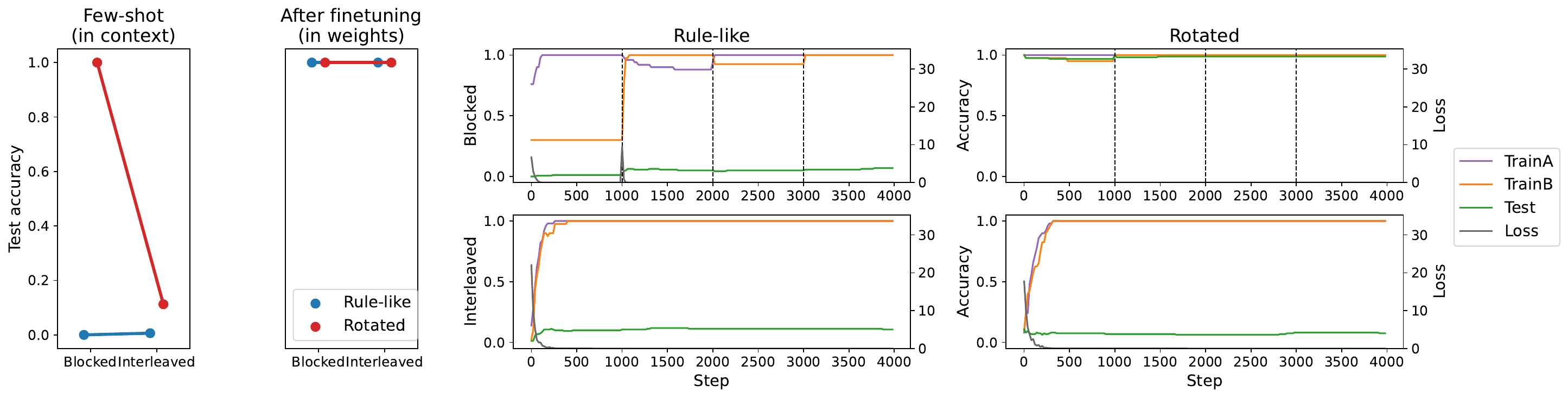}
\caption{Results for a metalearning network trained on a distribution of rotated tasks. This network acquires the ability to solve new tasks in-context, but shows the opposite curriculum effects as the networks trained on a distribution of rule-like tasks (compare to Figure 3g-i in the main text). In the few-shot evaluation, the network shows good performance on the rotated version of the task (red) and poor performance on the rule-like task (blue). Importantly, a blocking advantage was observed in the rotated, rather than the rule-like setting, consistent with our hypothesis that the blocking advantage emerges on tasks where ICL is possible. Task-specific training results show that IWL promotes catastrophic forgetting on the rule-like version of the task, rather than the rotated task. Interestingly, this catastrophic forgetting was not as severe compared to the network that saw rule-like tasks during metalearning and a rotated task during subsequent task-specific training (see Figure 3i in the main text), suggesting that the network was able to learn a strategy in the first block that did not interfere as much with the learning in the second block.}
\label{fig:rotated_results}
\end{figure}

What we have called the Rule-like versions of our tasks are only ``rule-like'' in the sense that the rules governing the task are simpler to discover because the features determining the correct answers (e.g., line orientation and length, or color and animal identities) are intuitive to humans. 
The rotated versions of the two tasks are more difficult because the rules of the task are only succinctly describable in \emph{rotated} feature spaces that are unintuitive.
Likewise, the rule-like versions of the tasks are easier for the metalearning networks to solve in-context because the appropriate features are familiar from the task distributions on which they metalearned. 
If instead the networks were more familiar with these rotated feature spaces, we would expect in-context learning to be \emph{easier} on tasks whose rules utilized similar feature spaces. 
To test this hypothesis, we trained metalearning networks on a distribution of rotated tasks in the compositional setting (see Figure \ref{fig:rotated_results}). 
As expected, the network showed the opposite pattern of curriculum effects, exhibiting a blocking advantage in the rotated task rather than the rule-like task. 
The network also suffered from catastrophic forgetting in the Rule-like task rather than the Rotated task.
Interestingly, catastrophic forgetting was not as severe compared to the network trained on a distribution of rule-like tasks before undergoing task-specific training on a rotated task. 
This meant that by the fourth block, the interleaving advantage observed through earlier blocks had disappeared.
This may suggest that metalearning on the rotated version of the task equipped the network with an inductive bias to learn strategies that suffer from less interference when related trials are blocked over time.

In general, this experiment is consistent with our hypothesis that the blocking advantage should emerge on tasks where the properties of ICL dominate, and the interleaving advantage should emerge on tasks where properties of IWL dominate due to catastrophic forgetting.
However, the task-specific training results suggest that catastrophic forgetting in IWL can be more or less severe depending on the inductive biases imparted by metalearning.

\subsection{Metalearning on interleaved curricula}
\begin{figure}[h]
\centering
\includegraphics[width=\textwidth]{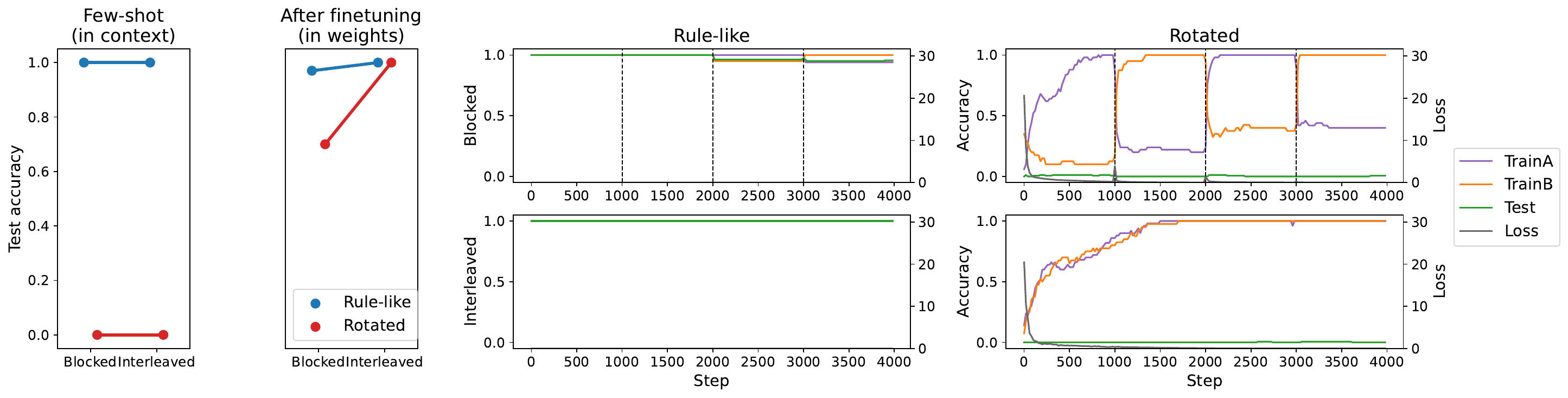}
\caption{Results for a metalearning network trained on a distribution of rule-like tasks with interleaved curricula. Due to the asymmetric nature of the blocking vs. interleaving distinction (a blocked curriculum can be seen as a special case of a randomly shuffled, or interleaved, curriculum) this network is capable of learning new tasks via ICL in either curriculum condition.}
\label{fig:interleaved_results}
\end{figure}

Our main results show that for both the category-learning task and the compositional task, curriculum effects can depend on whether ICL or IWL is dominant during learning: an interleaving advantage emerged whenever IWL was dominant (e.g., in the rotated versions of the tasks), and a blocking advantage emerged whenever ICL was dominant (e.g., in the rule-like versions of the tasks). 
While the interleaving advantage is an inherent property of IWL, the blocking advantage we observed in ICL emerged due to the use of blocked curricula during metalearning.
For the sake of completeness, here we report results for the compositional task in the case where interleaved curricula were used during metalearning.
One might predict that metalearning from interleaved curricula would impart an interleaving advantage, rather than a blocking advantage, to the resulting ICL algorithm.
However, it is important to note that if interleaving is defined as random shuffling, a blocked curriculum is a special case of an interleaved curriculum, where the individual examples happen to be organized into blocks.
This means that if a model has metalearned to perform well on all possible interleaved curricula, it will have learned an ICL algorithm that is robust to any changes in trial order, and should perform well even when trials are blocked.
Figure \ref{fig:interleaved_results} shows the results from a model after metalearning on interleaved curricula.
As expected, the model is capable of learning new tasks via ICL in both the blocked and interleaved curriculum conditions. 
These results suggest that in our setup, a blocking advantage is not an inherent property of ICL per se, but emerges due to the data distribution used during metalearning.

\subsection{Blocking over context vs. gradient steps}
The curriculum conditions in our experiments manipulate the order in which examples are presented over time. 
However, transformers afford multiple notions of time: they are given an entire sequence (which itself can be indexed by time/position) as input and are trained by backpropagating errors to incrementally update their weights for multiple gradient steps over training time. 
We can therefore differentiate whether related trials are blocked over the \emph{context} (i.e., related study examples are blocked over the context within a given input sequence) or blocked over the gradient \emph{steps} (i.e., sequences containing related test queries are used to compute and backpropagate losses in a blocked fashion over the gradient steps).
In our main experiments, we collapse these two notions by maintaining congruence between them: in the ``blocked'' condition, related trials were blocked over both the context and the gradient steps, and in the ``interleaved'' condition, trials were randomly interleaved over both the context and the gradient steps. 
These congruent cases are more realistic when compared with the human situation, where recently seen items that might be held in working memory (and therefore supply a context for the current trial) are the same ones that have recently been used to update synaptic weights (assuming that synapses are updating in a relatively continuous manner over time).

For completeness and to facilitate understanding of our main results, we also performed experiments where these two notions of time are incongruent (i.e., blocked over the context and interleaved over the gradient steps, or interleaved over the context and blocked over the gradient steps).
Here, we report such results from the same metalearning model trained on the compositional task (see Figure 3g-i in the main text). 
After metalearning, the model underwent task-specific training in the usual way but with these extra two incongruent curriculum conditions.

\begin{figure}[h]
\centering
\includegraphics[width=\textwidth]{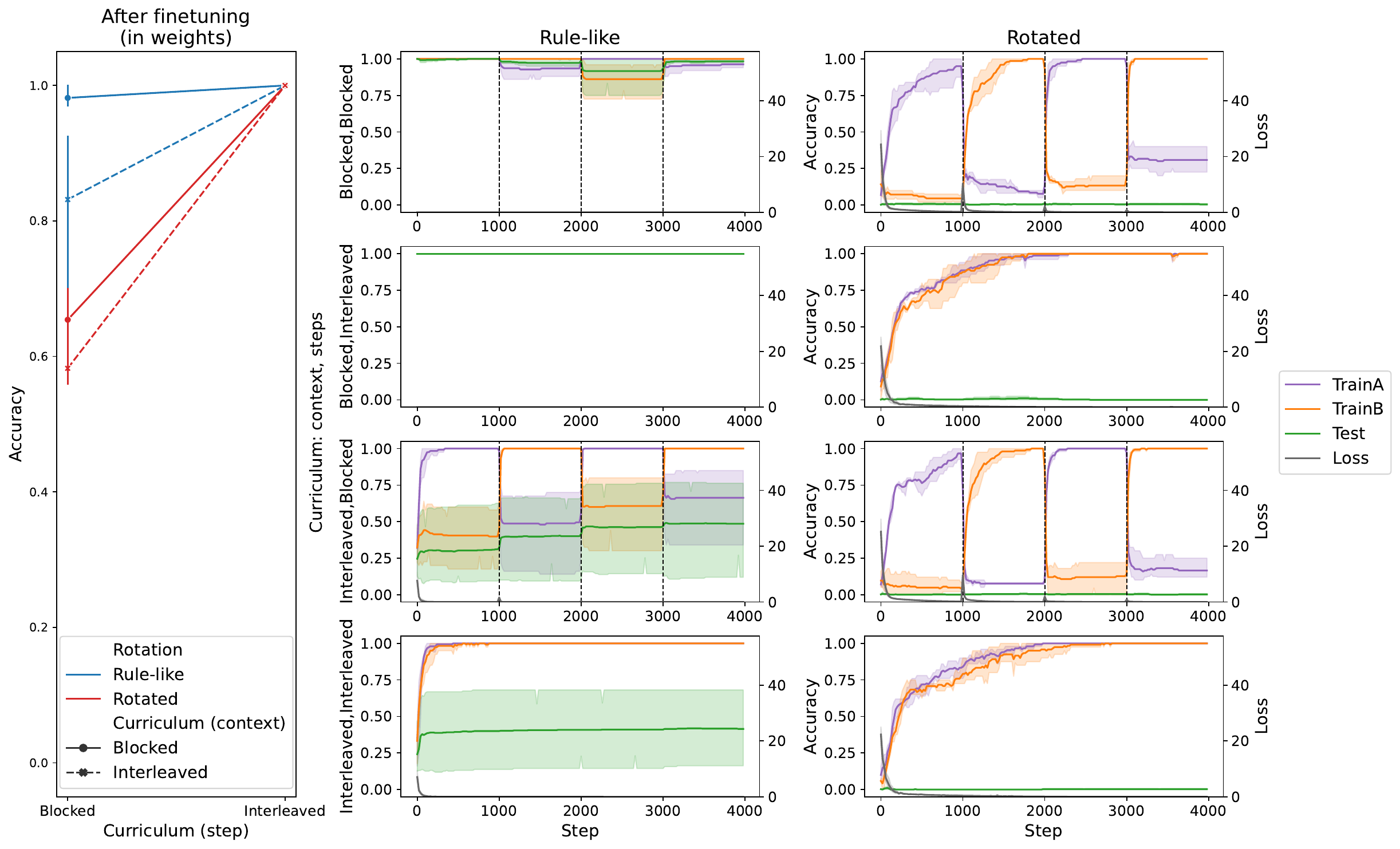}
\caption{Results of experiments investigating effects of blocking over the study examples given in the context vs. blocking over the gradient steps during task-specific training. The task-specific training results from Figure 3h-i in the main text report the main cases of interest where trials were either blocked over both context and gradient steps or interleaved over both context and gradient steps. Here, we show these same results alongside the cases where the two are mismatched. When trials are blocked over the context but interleaved over the steps (second row), the model shows good compositional generalization in the rule-like task and does not suffer from catastrophic forgetting. When trials are interleaved over the context but blocked over the steps (third row), the model shows poor compositional generalization and suffers from catastrophic forgetting. These results demonstrate how this model generalizes compositionally only when trials are blocked over the context in the rule-like task, and how catastrophic forgetting occurs whenever large losses are incurred (inducing IWL) and trials are blocked over the gradient steps.}
\label{fig:context_steps}
\end{figure}

The results are presented alongside the original (congruent) results in Figure \ref{fig:context_steps}.
In general, the results show that the key dynamics motivating our original theoretical framework hold across these incongruent conditions.
In the metalearning model, ICL is successful when encountering new tasks that have familiar structure (in this case, rule-like tasks where related items are blocked over the context).
This is shown by the high train (purple and orange) and test (green) accuracy, and the low loss (black) from the beginning of task-specific training in the rule-like task when items are blocked over the context (middle column, top two plots).
When ICL is successful, compositional generalization performance is good, and little loss is incurred and backpropagated, resulting in less IWL.
This can be seen by the lack of catastrophic forgetting in these same two plots.
On the other hand, when ICL is unsuccessful (in this case, when the task is rotated or when items are interleaved over the context), compositional generalization is poor, and large losses are backpropagated, resulting in increased IWL.
This can be seen in the loss curves (black) in the rotated task (right column), and in the rule-like task when items are interleaved over the context (middle column, bottom two plots). 
This increased IWL results in catastrophic forgetting when related trials are blocked over the gradient steps.
This can be seen in the sharp drops in accuracy on the items trained in the previous block (e.g., TrainA accuracy, shown in purple, drops during the second block as TrainB items are learned) in the rule-like task when items are interleaved over the context but blocked over the steps (middle column, third plot) or in the rotated task when items are blocked over the gradient steps (right column, first and third plots).
When trials are randomly interleaved over the gradient steps, no catastrophic forgetting occurs, even in cases where ICL is unsuccessful. 
This can be seen in the rule-like task when trials are interleaved over both the context and the gradient steps (middle column, fourth plot), and in the rotated task when items are interleaved over the gradient steps (right column, second and fourth plots). 

To summarize, these results show that compositional generalization performance is high when ICL is possible because the model is familiar with the structure of the task given in context (i.e. when the task is rule-like and related items are blocked over the context), and catastrophic forgetting happens when large losses are incurred (resulting in increased IWL) and related items are blocked over the gradient steps.
Our theoretical framework assumes that in humans, both contextual information and synaptic weights are updated more-or-less continuously throughout learning in the experimental tasks we model.
This implies that the distinction between these two notions of time, and the corresponding distinction between blocking/interleaving over the context vs. over the gradient steps, would collapse into a single notion.

\subsection{Tradeoff between flexibility and retention in category-learning task}
\begin{figure}[h]
\centering
\includegraphics[width=\textwidth]{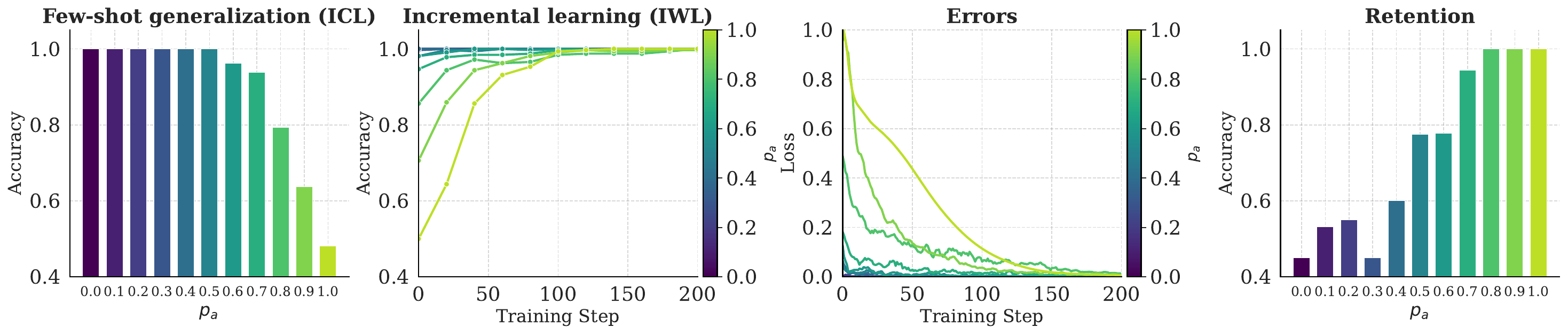}
\caption{Results of ablating attention in category-learning task. Greater ablation ($p_a$) resulted in worse few-shot generalization via ICL (left). When the model was not capable of solving the task in context, incremental learning via IWL took longer to reach optimal performance (middle left), and more errors were accumulated over the course of learning (middle right). These errors resulted in greater IWL and better retention when no contextual information was available (right). $p_a$ is the probability of inhibiting attention to each example in the context.}
\label{fig:tradeoff_category}
\end{figure}

In the main simulations investigating the tradeoff between flexibility and retention (see Figure 5 of the main text), we used the compositional task.
Here, we report the results of performing the same manipulation in the category-learning task (see Figure \ref{fig:tradeoff_category}).
Ablating attention during task-specific training had similar effects in the category-learning setting: greater ablations (larger $p_a$) resulted in worse few-shot generalization performance via ICL (left), and required more incremental learning via IWL for models to reach optimal performance (middle left). 
This meant that more errors were accumulated throughout task-specific training (middle right), resulting in more IWL and better subsequent retention (right). 

These results qualitatively reproduce the effects we observed on the compositional task (see Figure 5 of the main text), and are consistent with studies showing a similar tradeoff in human reinforcement learning \cite{CollinsFrank18b, Rac-LubashevskyCremerCollinsEtAl23}.

\subsection{Ablating attention by applying noise to value vectors}

The ablation simulations reported above and in the main text (see Figure 5) masked attention to individual examples in the context with probability $p_a$. 
We also experimented with a different kind of ablation, adding Gaussian noise to the attention mechanism itself.
This allowed us to replicate our previous ablation results while manipulating the model's ability to access to contextual information in a different way. 

\begin{figure}[h!]
\centering
\includegraphics[width=\textwidth]{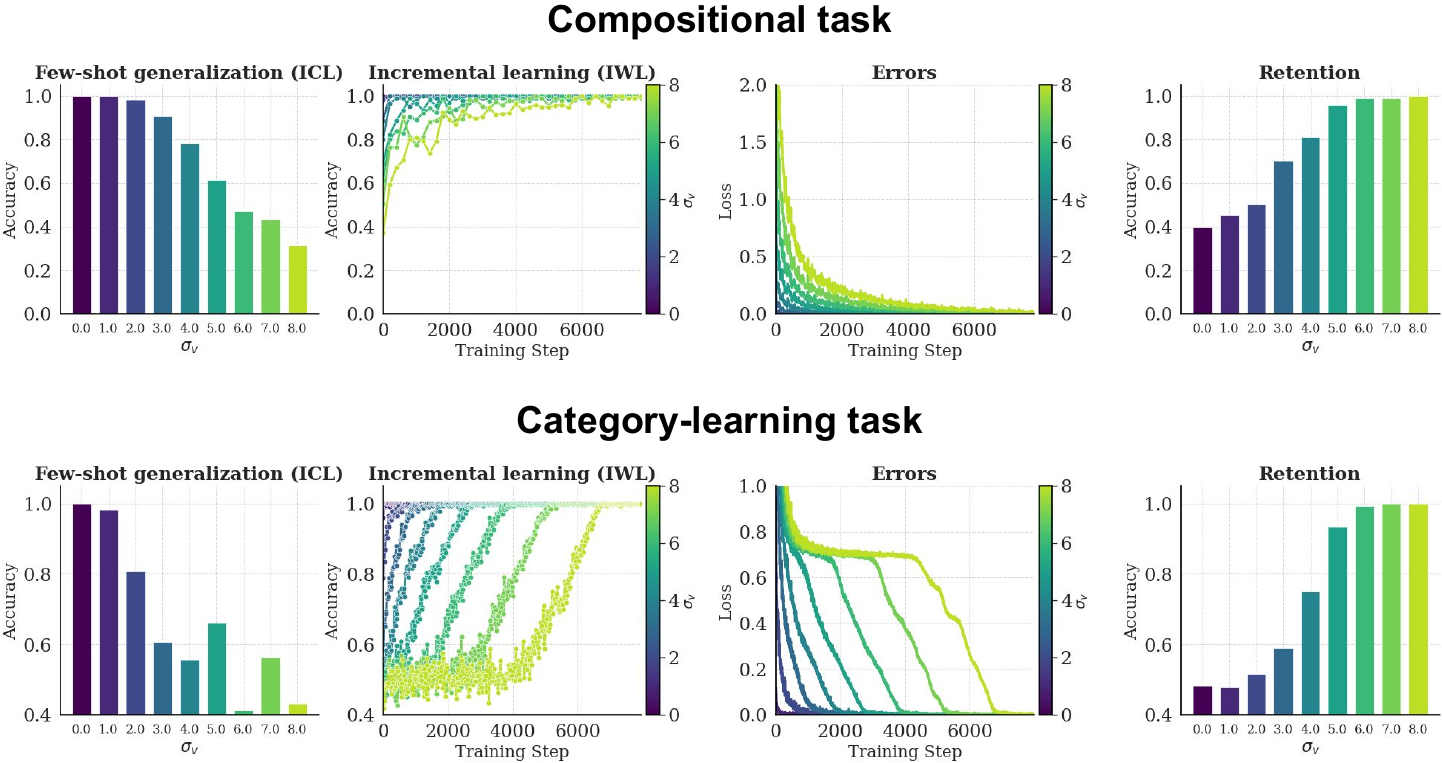}
\caption{Results of injecting noise into value vectors. Gaussian noise with standard deviation $\sigma_{v}$ was added to the value vectors across all tokens in the context. In both tasks, injecting noise in this way had effects similar to those observed when attention was masked, revealing the same tradeoff between flexibility and retention. More noise (high $\sigma_{v}$) resulted in worse flexibility (few-shot generalization via ICL; left), slower incremental learning via IWL (middle left), more errors (middle right), and greater retention (right).}
\label{fig:tradeoff_gaussian}
\end{figure}

We added Gaussian noise with standard deviation $\sigma_{v}$ to the value vectors corresponding to all tokens in the context.
This was accomplished by running a forward pass on the model, collecting the past keys and values, and applying the noise to the value vectors before running the forward pass again to produce predictions. 
We used the same experimental design as was used in the simulations that ablated attention using a mask (see above).
Again, models that metalearned on unrotated tasks with interleaved curricula were finetuned under different amounts of noise.
Few-shot generalization performance was measured with different levels of noise applied at inference time.
Different levels of noise were applied throughout finetuning, while recording training accuracy (incremental learning via IWL) and errors (i.e., loss).
Retention was tested by applying the maximum amount of noise ($\sigma_{v} = 8.0$) at inference time after finetuning.

The results are shown in Figure \ref{fig:tradeoff_gaussian}.
Again, we observed the same tradeoff between flexibility and retention exhibited by humans \cite{CollinsFrank18b, Rac-LubashevskyCremerCollinsEtAl23}.
In both tasks, noise degraded ICL few-shot generalization performance, but improved later retention.
These results replicate our original findings using a different manipulation, and highlight the general principle that restricting ICL hurts flexibility (few-shot generalization) and improves retention.

\newpage
\bibliographystyle{unsrt}
\bibliography{main}

\begin{thebibliography}{100}

\bibitem{AshbyMaddox11}
F.~Gregory Ashby and W.~Todd Maddox.
\newblock Human category learning 2.0.
\newblock {\em Annals of the New York Academy of Sciences}, 1224(1):147--161, 2011.

\bibitem{BotvinickRitterWangEtAl19a}
Matthew Botvinick, Sam Ritter, Jane~X. Wang, Zeb {Kurth-Nelson}, Charles Blundell, and Demis Hassabis.
\newblock Reinforcement {{Learning}}, {{Fast}} and {{Slow}}.
\newblock {\em Trends in Cognitive Sciences}, 23(5):408--422, May 2019.

\bibitem{DawNivDayan05}
Nathaniel~D. Daw, Yael Niv, and Peter Dayan.
\newblock Uncertainty-based competition between prefrontal and dorsolateral striatal systems for behavioral control.
\newblock {\em Nature Neuroscience}, 8(12):1704--1711, November 2005.

\bibitem{Evans08}
Jonathan St B.~T. Evans.
\newblock Dual-processing accounts of reasoning, judgment, and social cognition.
\newblock {\em Annual Review of Psychology}, 59:255--278, 2008.

\bibitem{NohYanBjorkEtAl16}
Sharon~M. Noh, Veronica~X. Yan, Robert~A. Bjork, and W.~Todd Maddox.
\newblock Optimal sequencing during category learning: {{Testing}} a dual-learning systems perspective.
\newblock {\em Cognition}, 155:23--29, 2016.

\bibitem{OReillyNairRussinEtAl20}
Randall~C. O'Reilly, Ananta Nair, Jacob~L. Russin, and Seth~A. Herd.
\newblock How {{Sequential Interactive Processing Within Frontostriatal Loops Supports}} a {{Continuum}} of {{Habitual}} to {{Controlled Processing}}.
\newblock {\em Frontiers in Psychology}, 11, 2020.

\bibitem{Sable-MeyerBenjaminWatkinsEtAl24}
Mathias {Sabl{\'e}-Meyer}, Lucas Benjamin, Cassandra~Potier Watkins, Chenxi He, Fosca~Al Roumi, and Stanislas Dehaene.
\newblock Two brain systems for the perception of geometric shapes, March 2024.

\bibitem{Sloman96a}
Steven~A. Sloman.
\newblock The empirical case for two systems of reasoning.
\newblock {\em Psychological Bulletin}, 119(1):3--22, 1996.

\bibitem{CollinsFrank18b}
Anne G.~E. Collins and Michael~J. Frank.
\newblock Within- and across-trial dynamics of human {{EEG}} reveal cooperative interplay between reinforcement learning and working memory.
\newblock {\em Proceedings of the National Academy of Sciences}, 115(10):2502--2507, March 2018.

\bibitem{CollinsFrank13a}
Anne G.~E. Collins and Michael~J. Frank.
\newblock Cognitive control over learning: {{Creating}}, clustering, and generalizing task-set structure.
\newblock {\em Psychological Review}, 120(1):190--229, January 2013.

\bibitem{CollinsFrank16b}
Anne Gabrielle~Eva Collins and Michael~Joshua Frank.
\newblock Neural signature of hierarchically structured expectations predicts clustering and transfer of rule sets in reinforcement learning.
\newblock {\em Cognition}, 152:160--169, July 2016.

\bibitem{LakeUllmanTenenbaumEtAl17a}
Brenden~M. Lake, Tomer~D. Ullman, Joshua~B. Tenenbaum, and Samuel~J. Gershman.
\newblock Building machines that learn and think like people.
\newblock {\em Behavioral and Brain Sciences}, 40, 2017/ed.

\bibitem{OReillyFrank06}
R.~C. O'Reilly and Michael~J. Frank.
\newblock Making working memory work: {{A}} computational model of learning in the prefrontal cortex and basal ganglia.
\newblock {\em Neural Computation}, 18(2):283--328, 2006.

\bibitem{RougierNoelleBraverEtAl05d}
Nicolas~P. Rougier, David~C. Noelle, Todd~S. Braver, Jonathan~D. Cohen, and Randall~C. O'Reilly.
\newblock Prefrontal cortex and flexible cognitive control: {{Rules}} without symbols.
\newblock {\em Proceedings of the National Academy of Sciences}, 102(20):7338--7343, May 2005.

\bibitem{FleschBalaguerDekkerEtAl18}
Timo Flesch, Jan Balaguer, Ronald Dekker, Hamed Nili, and Christopher Summerfield.
\newblock Comparing continual task learning in minds and machines.
\newblock {\em Proceedings of the National Academy of Sciences}, 115(44):E10313--E10322, October 2018.

\bibitem{PesnotLerousseauSummerfield24}
Jacques Pesnot~Lerousseau and Christopher Summerfield.
\newblock Space as a scaffold for rotational generalisation of abstract concepts.
\newblock {\em eLife}, 13:RP93636, April 2024.

\bibitem{Rac-LubashevskyCremerCollinsEtAl23}
Rachel {Rac-Lubashevsky}, Anna Cremer, Anne G.~E. Collins, Michael~J. Frank, and Lars Schwabe.
\newblock Neural {{Index}} of {{Reinforcement Learning Predicts Improved Stimulus}}--{{Response Retention}} under {{High Working Memory Load}}.
\newblock {\em Journal of Neuroscience}, 43(17):3131--3143, April 2023.

\bibitem{LakeSalakhutdinovGrossEtAl11}
Brenden Lake, Ruslan Salakhutdinov, Jason Gross, and Joshua Tenenbaum.
\newblock One shot learning of simple visual concepts.
\newblock {\em Proceedings of the Annual Meeting of the Cognitive Science Society}, 33(33), 2011.

\bibitem{LakeSalakhutdinovTenenbaum15b}
Brenden~M. Lake, Ruslan Salakhutdinov, and Joshua~B. Tenenbaum.
\newblock Human-level concept learning through probabilistic program induction.
\newblock {\em Science}, 350(6266):1332--1338, December 2015.

\bibitem{DekkerOttoSummerfield22c}
Ronald~B. Dekker, Fabian Otto, and Christopher Summerfield.
\newblock Curriculum learning for human compositional generalization.
\newblock {\em Proceedings of the National Academy of Sciences}, 119(41):e2205582119, October 2022.

\bibitem{FodorPylyshyn88a}
Jerry~A. Fodor and Zenon~W. Pylyshyn.
\newblock Connectionism and cognitive architecture: {{A}} critical analysis.
\newblock {\em Cognition}, 28(1-2):3--71, March 1988.

\bibitem{FranklinFrank18a}
Nicholas~T. Franklin and Michael~J. Frank.
\newblock Compositional clustering in task structure learning.
\newblock {\em PLOS Computational Biology}, 14(4):e1006116, April 2018.

\bibitem{FranklinFrank20a}
Nicholas~T. Franklin and Michael~J. Frank.
\newblock Generalizing to generalize: {{Humans}} flexibly switch between compositional and conjunctive structures during reinforcement learning.
\newblock {\em PLOS Computational Biology}, 16(4):e1007720, April 2020.

\bibitem{FranklandGreene20}
Steven~M. Frankland and Joshua~D. Greene.
\newblock Concepts and {{Compositionality}}: {{In Search}} of the {{Brain}}'s {{Language}} of {{Thought}}.
\newblock {\em Annual Review of Psychology}, 71(1):273--303, 2020.

\bibitem{LakeLinzenBaroni19a}
Brenden~M. Lake, Tal Linzen, and Marco Baroni.
\newblock Human few-shot learning of compositional instructions.
\newblock In Ashok~K. Goel, Colleen~M. Seifert, and Christian Freksa, editors, {\em Proceedings of the 41th {{Annual Meeting}} of the {{Cognitive Science Society}}, {{CogSci}} 2019: {{Creativity}} + {{Cognition}} + {{Computation}}, {{Montreal}}, {{Canada}}, {{July}} 24-27, 2019}, pages 611--617. cognitivesciencesociety.org, 2019.

\bibitem{LiuFrank22}
Rex~G. Liu and Michael~J. Frank.
\newblock Hierarchical clustering optimizes the tradeoff between compositionality and expressivity of task structures for flexible reinforcement learning.
\newblock {\em Artificial Intelligence}, 312:103770, November 2022.

\bibitem{RussinEtAl2024}
Jacob Russin, Sam~Whitman McGrath, Ellie Pavlick, and Michael~J. Frank.
\newblock Is human compositionality meta-learned?
\newblock {\em Commentary in Behavioral and Brain Sciences (forthcoming)}, 2024.

\bibitem{SchwartenbeckBaramLiuEtAl23}
Philipp Schwartenbeck, Alon Baram, Yunzhe Liu, Shirley Mark, Timothy Muller, Raymond Dolan, Matthew Botvinick, Zeb {Kurth-Nelson}, and Timothy Behrens.
\newblock Generative replay underlies compositional inference in the hippocampal-prefrontal circuit.
\newblock {\em Cell}, 186(22):4885--4897.e14, October 2023.

\bibitem{BeukersCollinKempnerEtAl24a}
Andre~O. Beukers, Silvy H.~P. Collin, Ross~P. Kempner, Nicholas~T. Franklin, Samuel~J. Gershman, and Kenneth~A. Norman.
\newblock Blocked training facilitates learning of multiple schemas.
\newblock {\em Communications Psychology}, 2(1):1--17, April 2024.

\bibitem{FrankBadre12}
Michael~J. Frank and David Badre.
\newblock Mechanisms of hierarchical reinforcement learning in corticostriatal circuits 1: {{Computational}} analysis.
\newblock {\em Cerebral Cortex (New York, N.Y.: 1991)}, 22(3):509--526, March 2012.

\bibitem{RichlandFinleyBjork04}
Lindsey~E. Richland, Jason~R. Finley, and Robert~A. Bjork.
\newblock Differentiating the {{Contextual Interference Effect}} from the {{Spacing Effect}}.
\newblock {\em Proceedings of the Annual Meeting of the Cognitive Science Society}, 26(26), 2004.

\bibitem{GoodeMagill86}
Sinah Goode and Richard~A. Magill.
\newblock Contextual {{Interference Effects}} in {{Learning Three Badminton Serves}}.
\newblock {\em Research Quarterly for Exercise and Sport}, 57(4):308--314, December 1986.

\bibitem{LandinHebertFairweather93}
Dennis~K. Landin, Edward~P. Hebert, and Malcolm Fairweather.
\newblock The {{Effects}} of {{Variable Practice}} on the {{Performance}} of a {{Basketball Skill}}.
\newblock {\em Research Quarterly for Exercise and Sport}, 64(2):232--237, June 1993.

\bibitem{Kahneman11}
Daniel Kahneman.
\newblock {\em Thinking, {{Fast}} and {{Slow}}}.
\newblock {Farrar, Straus and Giroux}, 1 edition edition, October 2011.

\bibitem{RumelhartHintonWilliams86}
David~E. Rumelhart, Geoffrey~E. Hinton, and Ronald~J. Williams.
\newblock Learning representations by back-propagating errors.
\newblock {\em Nature}, 323(6088):533--536, October 1986.

\bibitem{RumelhartMcClellandPDPResearchGroup86}
David~E. Rumelhart, James~L. McClelland, and PDP~Research Group, editors.
\newblock {\em Parallel Distributed Processing: Explorations in the Microstructure of Cognition, Vol. 2: Psychological and Biological Models}.
\newblock MIT Press, Cambridge, MA, USA, 1986.

\bibitem{LakeBaroni18}
Brenden~M. Lake and Marco Baroni.
\newblock Generalization without {{Systematicity}}: {{On}} the {{Compositional Skills}} of {{Sequence-to-Sequence Recurrent Networks}}.
\newblock In Jennifer~G. Dy and Andreas Krause, editors, {\em Proceedings of the 35th {{International Conference}} on {{Machine Learning}}, {{ICML}} 2018, {{Stockholmsm{\"a}ssan}}, {{Stockholm}}, {{Sweden}}, {{July}} 10-15, 2018}, volume~80 of {\em Proceedings of {{Machine Learning Research}}}, pages 2879--2888. PMLR, 2018.

\bibitem{Marcus98}
Gary~F. Marcus.
\newblock Rethinking {{Eliminative Connectionism}}.
\newblock {\em Cognitive Psychology}, 37(3):243--282, December 1998.

\bibitem{Marcus18}
Gary Marcus.
\newblock Deep learning: {{A}} critical appraisal, January 2018.

\bibitem{PinkerPrince88a}
Steven Pinker and Alan Prince.
\newblock On language and connectionism: {{Analysis}} of a parallel distributed processing model of language acquisition.
\newblock {\em Cognition}, 28(1):73--193, March 1988.

\bibitem{Quilty-DunnPorotMandelbaum23}
Jake {Quilty-Dunn}, Nicolas Porot, and Eric Mandelbaum.
\newblock The best game in town: {{The}} reemergence of the language-of-thought hypothesis across the cognitive sciences.
\newblock {\em Behavioral and Brain Sciences}, 46:e261, January 2023.

\bibitem{RussinZolfagharParkEtAl22}
Jacob Russin, Maryam Zolfaghar, Seongmin~A. Park, Erie Boorman, and Randall~C. O'Reilly.
\newblock A {{Neural Network Model}} of {{Continual Learning}} with {{Cognitive Control}}.
\newblock In {\em Proceedings for the 44th {{Annual Meeting}} of the {{Cognitive Science Society}}}, February 2022.

\bibitem{McCloskeyCohen89}
M.~McCloskey and N.~J. Cohen.
\newblock Catastrophic {{Interference}} in {{Connectionist Networks}}: {{The Sequential Learning Problem}}.
\newblock In G.~H. Bower, editor, {\em The {{Psychology}} of {{Learning}} and {{Motivation}}, {{Vol}}. 24}, pages 109--164. Academic Press, San Diego, CA, January 1989.

\bibitem{McClellandMcNaughtonOReilly95}
J.~L. McClelland, B.~L. McNaughton, and R.~C. O'Reilly.
\newblock Why {{There Are Complementary Learning Systems}} in the {{Hippocampus}} and {{Neocortex}}: {{Insights}} from the {{Successes}} and {{Failures}} of {{Connectionist Models}} of {{Learning}} and {{Memory}}.
\newblock {\em Psychological Review}, 102(3):419--457, August 1995.

\bibitem{LoveMedinGureckis04a}
Bradley~C. Love, Douglas~L. Medin, and Todd~M. Gureckis.
\newblock {{SUSTAIN}}: {{A Network Model}} of {{Category Learning}}.
\newblock {\em Psychological Review}, 111(2):309--332, 2004.

\bibitem{FrankClaus06}
Michael~J. Frank and Eric~D. Claus.
\newblock Anatomy of a decision: {{Striato-orbitofrontal}} interactions in reinforcement learning, decision making, and reversal.
\newblock {\em Psychological Review}, 113(2):300--326, April 2006.

\bibitem{KrieteNoelleCohenEtAl13}
T.~Kriete, D.~C. Noelle, J.~D. Cohen, and R.~C. O'Reilly.
\newblock Indirection and symbol-like processing in the prefrontal cortex and basal ganglia.
\newblock {\em Proceedings of the National Academy of Sciences}, 110(41):16390--16395, October 2013.

\bibitem{MillerCohen01}
E.~K. Miller and J.~D. Cohen.
\newblock An integrative theory of prefrontal cortex function.
\newblock {\em Annual Review of Neuroscience}, 24:167--202, 2001.

\bibitem{WangKurth-NelsonKumaranEtAl18a}
Jane~X. Wang, Zeb {Kurth-Nelson}, Dharshan Kumaran, Dhruva Tirumala, Hubert Soyer, Joel~Z. Leibo, Demis Hassabis, and Matthew Botvinick.
\newblock Prefrontal cortex as a meta-reinforcement learning system.
\newblock {\em Nature Neuroscience}, 21(6):860--868, June 2018.

\bibitem{BrownMannRyderEtAl20b}
Tom~B. Brown, Benjamin Mann, Nick Ryder, Melanie Subbiah, Jared Kaplan, Prafulla Dhariwal, Arvind Neelakantan, Pranav Shyam, Girish Sastry, Amanda Askell, Sandhini Agarwal, Ariel {Herbert-Voss}, Gretchen Krueger, Tom Henighan, Rewon Child, Aditya Ramesh, Daniel~M. Ziegler, Jeffrey Wu, Clemens Winter, Christopher Hesse, Mark Chen, Eric Sigler, Mateusz Litwin, Scott Gray, Benjamin Chess, Jack Clark, Christopher Berner, Sam McCandlish, Alec Radford, Ilya Sutskever, and Dario Amodei.
\newblock Language {{Models}} are {{Few-Shot Learners}}, May 2020.

\bibitem{BubeckChandrasekaranEldanEtAl23}
S{\'e}bastien Bubeck, Varun Chandrasekaran, Ronen Eldan, Johannes Gehrke, Eric Horvitz, Ece Kamar, Peter Lee, Yin~Tat Lee, Yuanzhi Li, Scott Lundberg, Harsha Nori, Hamid Palangi, Marco~Tulio Ribeiro, and Yi~Zhang.
\newblock Sparks of {{Artificial General Intelligence}}: {{Early}} experiments with {{GPT-4}}, March 2023.

\bibitem{SaparovPangPadmakumarEtAl23}
Abulhair Saparov, Richard~Yuanzhe Pang, Vishakh Padmakumar, Nitish Joshi, Seyed~Mehran Kazemi, Najoung Kim, and He~He.
\newblock Testing the {{General Deductive Reasoning Capacity}} of {{Large Language Models Using OOD Examples}}, November 2023.

\bibitem{MuskerDuchnowskiMilliereEtAl24}
Sam Musker, Alex Duchnowski, Rapha{\"e}l Milli{\`e}re, and Ellie Pavlick.
\newblock Semantic {{Structure-Mapping}} in {{LLM}} and {{Human Analogical Reasoning}}, June 2024.

\bibitem{WebbHolyoakLu23}
Taylor Webb, Keith~J. Holyoak, and Hongjing Lu.
\newblock Emergent analogical reasoning in large language models.
\newblock {\em Nature Human Behaviour}, 7(9):1526--1541, September 2023.

\bibitem{LakeBaroni23a}
Brenden~M. Lake and Marco Baroni.
\newblock Human-like systematic generalization through a meta-learning neural network.
\newblock {\em Nature}, 623(7985):115--121, November 2023.

\bibitem{PressZhangMinEtAl23a}
Ofir Press, Muru Zhang, Sewon Min, Ludwig Schmidt, Noah~A. Smith, and Mike Lewis.
\newblock Measuring and {{Narrowing}} the {{Compositionality Gap}} in {{Language Models}}, October 2023.

\bibitem{ZhouScharliHouEtAl22a}
Denny Zhou, Nathanael Sch{\"a}rli, Le~Hou, Jason Wei, Nathan Scales, Xuezhi Wang, Dale Schuurmans, Claire Cui, Olivier Bousquet, Quoc~V. Le, and Ed~H. Chi.
\newblock Least-to-{{Most Prompting Enables Complex Reasoning}} in {{Large Language Models}}.
\newblock In {\em The {{Eleventh International Conference}} on {{Learning Representations}}}, September 2022.

\bibitem{ChanSantoroLampinenEtAl22}
Stephanie C.~Y. Chan, Adam Santoro, Andrew~K. Lampinen, Jane~X. Wang, Aaditya Singh, Pierre~H. Richemond, Jay McClelland, and Felix Hill.
\newblock Data {{Distributional Properties Drive Emergent In-Context Learning}} in {{Transformers}}, May 2022.

\bibitem{vonOswaldNiklassonSchlegelEtAl23}
Johannes {von Oswald}, Eyvind Niklasson, Maximilian Schlegel, Seijin Kobayashi, Nicolas Zucchet, Nino Scherrer, Nolan Miller, Mark Sandler, Blaise~Ag{\"u}era y~Arcas, Max Vladymyrov, Razvan Pascanu, and Jo{\~a}o Sacramento.
\newblock Uncovering mesa-optimization algorithms in {{Transformers}}, September 2023.

\bibitem{XieRaghunathanLiangEtAl22}
Sang~Michael Xie, Aditi Raghunathan, Percy Liang, and Tengyu Ma.
\newblock An {{Explanation}} of {{In-context Learning}} as {{Implicit Bayesian Inference}}, July 2022.

\bibitem{WangKurth-NelsonSoyerEtAl17}
Jane Wang, Zeb {Kurth-Nelson}, Hubert Soyer, Joel Leibo, Dhruva Tirumala, Remi Munos, Charles Blundell, Dharshan Kumaran, and Matt Botvinick.
\newblock Learning to reinforcement learn.
\newblock {\em Proceedings of the Annual Meeting of the Cognitive Science Society}, 39(0), 2017.

\bibitem{AnandLeporiMerulloEtAl24}
Suraj Anand, Michael~A. Lepori, Jack Merullo, and Ellie Pavlick.
\newblock Dual {{Process Learning}}: {{Controlling Use}} of {{In-Context}} vs. {{In-Weights Strategies}} with {{Weight Forgetting}}, May 2024.

\bibitem{Reddy23a}
Gautam Reddy.
\newblock The mechanistic basis of data dependence and abrupt learning in an in-context classification task, December 2023.

\bibitem{CollinsCiulloFrankEtAl17}
Anne~G.E. Collins, Brittany Ciullo, Michael~J. Frank, and David Badre.
\newblock Working {{Memory Load Strengthens Reward Prediction Errors}}.
\newblock {\em The Journal of Neuroscience}, 37(16):4332--4342, April 2017.

\bibitem{HitchcockKimFrank24}
Peter Hitchcock, Joonhwa Kim, and Michael Frank.
\newblock Working {{Memory}} and {{Reinforcement Learning Interactions}} when {{Simultaneously Pursuing Reward}} and {{Avoiding Punishment}}: {{No Relationship}} to {{Internalizing Symptoms}}, October 2024.

\bibitem{BinzDasguptaJagadishEtAl23a}
Marcel Binz, Ishita Dasgupta, Akshay~K. Jagadish, Matthew Botvinick, Jane~X. Wang, and Eric Schulz.
\newblock Meta-{{Learned Models}} of {{Cognition}}.
\newblock {\em Behavioral and Brain Sciences}, pages 1--38, November 2023.

\bibitem{FinnAbbeelLevine17d}
Chelsea Finn, Pieter Abbeel, and Sergey Levine.
\newblock Model-agnostic meta-learning for fast adaptation of deep networks.
\newblock In {\em Proceedings of the 34th {{International Conference}} on {{Machine Learning}} - {{Volume}} 70}, {{ICML}}'17, pages 1126--1135, Sydney, NSW, Australia, August 2017. JMLR.org.

\bibitem{JagadishCoda-FornoThalmannEtAl24}
Akshay~K. Jagadish, Julian {Coda-Forno}, Mirko Thalmann, Eric Schulz, and Marcel Binz.
\newblock Human-like {{Category Learning}} by {{Injecting Ecological Priors}} from {{Large Language Models}} into {{Neural Networks}}, May 2024.

\bibitem{SantoroBartunovBotvinickEtAl16}
Adam Santoro, Sergey Bartunov, Matthew Botvinick, Daan Wierstra, and Timothy Lillicrap.
\newblock Meta-{{Learning}} with {{Memory-Augmented Neural Networks}}, 2016.

\bibitem{RussinMcGrathEtAl2024}
Jacob Russin, Sam~Whitman McGrath, Danielle Williams, and Lotem {Elber-Dorozko}.
\newblock From {{Frege}} to {{chatGPT}}: {{Compositionality}} in language, cognition, and deep neural networks.
\newblock In {\em Forthcoming}. 2024.

\bibitem{KeysersScharliScalesEtAl19}
Daniel Keysers, Nathanael Sch{\"a}rli, Nathan Scales, Hylke Buisman, Daniel Furrer, Sergii Kashubin, Nikola Momchev, Danila Sinopalnikov, Lukasz Stafiniak, Tibor Tihon, Dmitry Tsarkov, Xiao Wang, Marc van Zee, and Olivier Bousquet.
\newblock Measuring {{Compositional Generalization}}: {{A Comprehensive Method}} on {{Realistic Data}}.
\newblock In {\em International {{Conference}} on {{Learning Representations}}}, September 2019.

\bibitem{KimLinzen20a}
Najoung Kim and Tal Linzen.
\newblock {{COGS}}: {{A Compositional Generalization Challenge Based}} on {{Semantic Interpretation}}.
\newblock In {\em Proceedings of the 2020 {{Conference}} on {{Empirical Methods}} in {{Natural Language Processing}} ({{EMNLP}})}, pages 9087--9105, Online, November 2020. Association for Computational Linguistics.

\bibitem{Frank23a}
Michael~C. Frank.
\newblock Bridging the data gap between children and large language models.
\newblock {\em Trends in Cognitive Sciences}, 0(0), August 2023.

\bibitem{Linzen20}
Tal Linzen.
\newblock How {{Can We Accelerate Progress Towards Human-like Linguistic Generalization}}?
\newblock In {\em Proceedings of the 58th {{Annual Meeting}} of the {{Association}} for {{Computational Linguistics}}}, pages 5210--5217, Online, 2020. Association for Computational Linguistics.

\bibitem{WarstadtMuellerChoshenEtAl23}
Alex Warstadt, Aaron Mueller, Leshem Choshen, Ethan Wilcox, Chengxu Zhuang, Juan Ciro, Rafael Mosquera, Bhargavi Paranjabe, Adina Williams, Tal Linzen, and Ryan Cotterell.
\newblock Findings of the {{BabyLM Challenge}}: {{Sample-Efficient Pretraining}} on {{Developmentally Plausible Corpora}}.
\newblock In Alex Warstadt, Aaron Mueller, Leshem Choshen, Ethan Wilcox, Chengxu Zhuang, Juan Ciro, Rafael Mosquera, Bhargavi Paranjabe, Adina Williams, Tal Linzen, and Ryan Cotterell, editors, {\em Proceedings of the {{BabyLM Challenge}} at the 27th {{Conference}} on {{Computational Natural Language Learning}}}, pages 1--34, Singapore, December 2023. Association for Computational Linguistics.

\bibitem{ChenChiWangEtAl24}
Xinyun Chen, Ryan~A. Chi, Xuezhi Wang, and Denny Zhou.
\newblock Premise {{Order Matters}} in {{Reasoning}} with {{Large Language Models}}, May 2024.

\bibitem{LuBartoloMooreEtAl22}
Yao Lu, Max Bartolo, Alastair Moore, Sebastian Riedel, and Pontus Stenetorp.
\newblock Fantastically {{Ordered Prompts}} and {{Where}} to {{Find Them}}: {{Overcoming Few-Shot Prompt Order Sensitivity}}.
\newblock In Smaranda Muresan, Preslav Nakov, and Aline Villavicencio, editors, {\em Proceedings of the 60th {{Annual Meeting}} of the {{Association}} for {{Computational Linguistics}} ({{Volume}} 1: {{Long Papers}})}, pages 8086--8098, Dublin, Ireland, May 2022. Association for Computational Linguistics.

\bibitem{TouvronMartinStoneEtAl23}
Hugo Touvron, Louis Martin, Kevin Stone, Peter Albert, Amjad Almahairi, Yasmine Babaei, Nikolay Bashlykov, Soumya Batra, Prajjwal Bhargava, Shruti Bhosale, Dan Bikel, Lukas Blecher, Cristian~Canton Ferrer, Moya Chen, Guillem Cucurull, David Esiobu, Jude Fernandes, Jeremy Fu, Wenyin Fu, Brian Fuller, Cynthia Gao, Vedanuj Goswami, Naman Goyal, Anthony Hartshorn, Saghar Hosseini, Rui Hou, Hakan Inan, Marcin Kardas, Viktor Kerkez, Madian Khabsa, Isabel Kloumann, Artem Korenev, Punit~Singh Koura, Marie-Anne Lachaux, Thibaut Lavril, Jenya Lee, Diana Liskovich, Yinghai Lu, Yuning Mao, Xavier Martinet, Todor Mihaylov, Pushkar Mishra, Igor Molybog, Yixin Nie, Andrew Poulton, Jeremy Reizenstein, Rashi Rungta, Kalyan Saladi, Alan Schelten, Ruan Silva, Eric~Michael Smith, Ranjan Subramanian, Xiaoqing~Ellen Tan, Binh Tang, Ross Taylor, Adina Williams, Jian~Xiang Kuan, Puxin Xu, Zheng Yan, Iliyan Zarov, Yuchen Zhang, Angela Fan, Melanie Kambadur, Sharan Narang, Aurelien Rodriguez, Robert Stojnic, Sergey Edunov, and Thomas
  Scialom.
\newblock Llama 2: {{Open Foundation}} and {{Fine-Tuned Chat Models}}, July 2023.

\bibitem{OuyangWuJiangEtAl22}
Long Ouyang, Jeff Wu, Xu~Jiang, Diogo Almeida, Carroll~L. Wainwright, Pamela Mishkin, Chong Zhang, Sandhini Agarwal, Katarina Slama, Alex Ray, John Schulman, Jacob Hilton, Fraser Kelton, Luke Miller, Maddie Simens, Amanda Askell, Peter Welinder, Paul Christiano, Jan Leike, and Ryan Lowe.
\newblock Training language models to follow instructions with human feedback, March 2022.

\bibitem{ChanDasguptaKimEtAl22a}
Stephanie C.~Y. Chan, Ishita Dasgupta, Junkyung Kim, Dharshan Kumaran, Andrew~K. Lampinen, and Felix Hill.
\newblock Transformers generalize differently from information stored in context vs in weights, October 2022.

\bibitem{Collins18}
Anne G.~E. Collins.
\newblock The {{Tortoise}} and the {{Hare}}: {{Interactions}} between {{Reinforcement Learning}} and {{Working Memory}}.
\newblock {\em Journal of Cognitive Neuroscience}, 30(10):1422--1432, October 2018.

\bibitem{StBTEvansStanovich13}
Jonathan Evans and Keith~E. Stanovich.
\newblock Dual-{{Process Theories}} of {{Higher Cognition}}: {{Advancing}} the {{Debate}}.
\newblock {\em Perspectives on psychological science}, 8(3):223--241, 2013.

\bibitem{ShiffrinSchneider77}
Richard~M. Shiffrin and Walter Schneider.
\newblock Controlled and automatic human information processing: {{II}}. {{Perceptual}} learning, automatic attending and a general theory.
\newblock {\em Psychological Review}, 84(2):127--190, 1977.

\bibitem{FabioCapriRomano19}
Rosa~Angela Fabio, Tindara Capr{\`i}, and Martina Romano.
\newblock From {{Controlled}} to {{Automatic Processes}} and {{Back Again}}: {{The Role}} of {{Contextual Features}}.
\newblock {\em Europe's Journal of Psychology}, 15(4):773--788, December 2019.

\bibitem{DawGershmanSeymourEtAl11}
Nathaniel~D. Daw, Samuel~J. Gershman, Ben Seymour, Peter Dayan, and Raymond~J. Dolan.
\newblock Model-based influences on humans' choices and striatal prediction errors.
\newblock {\em Neuron}, 69(6):1204--1215, March 2011.

\bibitem{SuttonBarto98}
R.~S. Sutton and A.~G. Barto.
\newblock {\em Reinforcement {{Learning}}: {{An Introduction}}.}
\newblock MIT Press, Cambridge, MA, January 1998.

\bibitem{AshbyMaddox05a}
F.~Gregory Ashby and W.~Todd Maddox.
\newblock Human {{Category Learning}}.
\newblock {\em Annual Review of Psychology}, 56(Volume 56, 2005):149--178, February 2005.

\bibitem{DongLiDaiEtAl24}
Qingxiu Dong, Lei Li, Damai Dai, Ce~Zheng, Jingyuan Ma, Rui Li, Heming Xia, Jingjing Xu, Zhiyong Wu, Tianyu Liu, Baobao Chang, Xu~Sun, Lei Li, and Zhifang Sui.
\newblock A {{Survey}} on {{In-context Learning}}, October 2024.

\bibitem{GrazziSiemsSchrodiEtAl24}
Riccardo Grazzi, Julien Siems, Simon Schrodi, Thomas Brox, and Frank Hutter.
\newblock Is {{Mamba Capable}} of {{In-Context Learning}}?, April 2024.

\bibitem{SushmaTianMesthaEtAl24}
Neeraj~Mohan Sushma, Yudou Tian, Harshvardhan Mestha, Nicolo Colombo, David Kappel, and Anand Subramoney.
\newblock State-space models can learn in-context by gradient descent, October 2024.

\bibitem{ZucchetKobayashiAkramEtAl24}
Nicolas Zucchet, Seijin Kobayashi, Yassir Akram, Johannes von Oswald, Maxime Larcher, Angelika Steger, and Jo{\~a}o Sacramento.
\newblock Gated recurrent neural networks discover attention, February 2024.

\bibitem{AkyurekSchuurmansAndreasEtAl23}
Ekin Aky{\"u}rek, Dale Schuurmans, Jacob Andreas, Tengyu Ma, and Denny Zhou.
\newblock What learning algorithm is in-context learning? {{Investigations}} with linear models, May 2023.

\bibitem{vonOswaldNiklassonRandazzoEtAl23}
Johannes {von Oswald}, Eyvind Niklasson, Ettore Randazzo, Jo{\~a}o Sacramento, Alexander Mordvintsev, Andrey Zhmoginov, and Max Vladymyrov.
\newblock Transformers learn in-context by gradient descent, May 2023.

\bibitem{OlssonElhageNandaEtAl22}
Catherine Olsson, Nelson Elhage, Neel Nanda, Nicholas Joseph, Nova DasSarma, Tom Henighan, Ben Mann, Amanda Askell, Yuntao Bai, Anna Chen, Tom Conerly, Dawn Drain, Deep Ganguli, Zac {Hatfield-Dodds}, Danny Hernandez, Scott Johnston, Andy Jones, Jackson Kernion, Liane Lovitt, Kamal Ndousse, Dario Amodei, Tom Brown, Jack Clark, Jared Kaplan, Sam McCandlish, and Chris Olah.
\newblock In-context {{Learning}} and {{Induction Heads}}, September 2022.

\bibitem{HendelGevaGloberson23}
Roee Hendel, Mor Geva, and Amir Globerson.
\newblock In-{{Context Learning Creates Task Vectors}}, October 2023.

\bibitem{ToddLiSharmaEtAl23}
Eric Todd, Millicent~L. Li, Arnab~Sen Sharma, Aaron Mueller, Byron~C. Wallace, and David Bau.
\newblock Function {{Vectors}} in {{Large Language Models}}, October 2023.

\bibitem{CarvalhoGoldstone14b}
Paulo~F. Carvalho and Robert~L. Goldstone.
\newblock Putting category learning in order: {{Category}} structure and temporal arrangement affect the benefit of interleaved over blocked study.
\newblock {\em Memory \& Cognition}, 42(3):481--495, April 2014.

\bibitem{OReillyBhattacharyyaHowardEtAl14a}
Randall~C. O'Reilly, Rajan Bhattacharyya, Michael~D. Howard, and Nicholas Ketz.
\newblock Complementary {{Learning Systems}}.
\newblock {\em Cognitive Science}, 38(6):1229--1248, 2014.

\bibitem{ParkMillerNiliEtAl20}
Seongmin~A. Park, Douglas~S. Miller, Hamed Nili, Charan Ranganath, and Erie~D. Boorman.
\newblock Map {{Making}}: {{Constructing}}, {{Combining}}, and {{Inferring}} on {{Abstract Cognitive Maps}}.
\newblock {\em Neuron}, 107(6):1226--1238.e8, 2020.

\bibitem{FleschNagySaxeEtAl23}
Timo Flesch, David~G. Nagy, Andrew Saxe, and Christopher Summerfield.
\newblock Modelling continual learning in humans with {{Hebbian}} context gating and exponentially decaying task signals.
\newblock {\em PLOS Computational Biology}, 19(1):e1010808, January 2023.

\bibitem{GiallanzaCampbellCohen24a}
Tyler Giallanza, Declan Campbell, and Jonathan~D. Cohen.
\newblock Toward the {{Emergence}} of {{Intelligent Control}}: {{Episodic Generalization}} and {{Optimization}}.
\newblock {\em Open Mind}, 8:688--722, May 2024.

\bibitem{SandbrinkSummerfield24}
Kai Sandbrink and Christopher Summerfield.
\newblock Modelling cognitive flexibility with deep neural networks.
\newblock {\em Current Opinion in Behavioral Sciences}, 57:101361, June 2024.

\bibitem{Wang21}
Jane~X Wang.
\newblock Meta-learning in natural and artificial intelligence.
\newblock {\em Current Opinion in Behavioral Sciences}, 38:90--95, April 2021.

\bibitem{MirchandaniXiaFlorenceEtAl23a}
Suvir Mirchandani, Fei Xia, Pete Florence, Brian Ichter, Danny Driess, Montserrat~Gonzalez Arenas, Kanishka Rao, Dorsa Sadigh, and Andy Zeng.
\newblock Large {{Language Models}} as {{General Pattern Machines}}, October 2023.

\bibitem{VaswaniShazeerParmarEtAl17b}
Ashish Vaswani, Noam Shazeer, Niki Parmar, Jakob Uszkoreit, Llion Jones, Aidan~N. Gomez, Lukasz Kaiser, and Illia Polosukhin.
\newblock Attention is all you need.
\newblock In Isabelle Guyon, Ulrike {von Luxburg}, Samy Bengio, Hanna~M. Wallach, Rob Fergus, S.~V.~N. Vishwanathan, and Roman Garnett, editors, {\em Adv. {{Neur}}. {{Inf}}. {{Proc}}. {{Sys}}. 30}, pages 5998--6008, long beach, CA, USA, 2017.

\bibitem{XuFutrell25}
Weijie Xu and Richard Futrell.
\newblock Strategic resource allocation in memory encoding: {{An}} efficiency principle shaping language processing, March 2025.

\bibitem{OReillyRanganathRussin22}
Randall~C. O'Reilly, Charan Ranganath, and Jacob~L. Russin.
\newblock The {{Structure}} of {{Systematicity}} in the {{Brain}}.
\newblock {\em Current directions in psychological science}, 31(2):124--130, April 2022.

\bibitem{PiantadosiAslin16}
Steven Piantadosi and Richard Aslin.
\newblock Compositional {{Reasoning}} in {{Early Childhood}}.
\newblock {\em PLOS ONE}, 11(9):e0147734, September 2016.

\bibitem{PiantadosiPalmeriAslin18}
Steven~T. Piantadosi, Holly Palmeri, and Richard Aslin.
\newblock Limits on composition of conceptual operations in 9-month-olds.
\newblock {\em Infancy : the official journal of the International Society on Infant Studies}, 23(3):310--324, 2018.

\bibitem{AshbyAlfonso-ReeseTurkenEtAl98}
F.~G. Ashby, L.~A. {Alfonso-Reese}, A.~U. Turken, and E.~M. Waldron.
\newblock A neuropsychological theory of multiple systems in category learning.
\newblock {\em Psychological Review}, 105(3):442--481, July 1998.

\bibitem{AshbySmithRosedahl20}
F.~Gregory Ashby, J.~David Smith, and Luke~A. Rosedahl.
\newblock Dissociations between rule-based and information-integration categorization are not caused by differences in task difficulty.
\newblock {\em Memory \& Cognition}, 48(4):541--552, May 2020.

\bibitem{NosofskyJohansen00}
R.~M. Nosofsky and M.~K. Johansen.
\newblock Exemplar-based accounts of "multiple-system" phenomena in perceptual categorization.
\newblock {\em Psychonomic Bulletin \& Review}, 7(3):375--402, September 2000.

\bibitem{Nosofsky11}
Robert~M. Nosofsky.
\newblock The generalized context model: {{An}} exemplar model of classification.
\newblock In {\em Formal Approaches in Categorization}, pages 18--39. Cambridge University Press, New York, NY, US, 2011.

\bibitem{MindaRoarkKalraEtAl24}
John~Paul Minda, Casey~L. Roark, Priya Kalra, and Anthony Cruz.
\newblock Single and multiple systems in categorization and category learning.
\newblock {\em Nature Reviews Psychology}, 3(8):536--551, August 2024.

\bibitem{NewellDunnKalish11a}
Ben~R. Newell, John~C. Dunn, and Michael Kalish.
\newblock Systems of category learning: {{Fact}} or fantasy?
\newblock In {\em The Psychology of Learning and Motivation: {{Advances}} in Research and Theory, {{Vol}}. 54}, The Psychology of Learning and Motivation, pages 167--215. Elsevier Academic Press, San Diego, CA, US, 2011.

\bibitem{PoldrackFoerde08}
Russell~A. Poldrack and Karin Foerde.
\newblock Category learning and the memory systems debate.
\newblock {\em Neuroscience \& Biobehavioral Reviews}, 32(2):197--205, January 2008.

\bibitem{StantonNosofsky07}
Roger~D. Stanton and Robert~M. Nosofsky.
\newblock Feedback interference and dissociations of classification: {{Evidence}} against the multiple-learning-systems hypothesis.
\newblock {\em Memory \& Cognition}, 35(7):1747--1758, October 2007.

\bibitem{MaddoxFiloteoHejlEtAl04}
W.~Todd Maddox, J.~Vincent Filoteo, Kelli~D. Hejl, and A.~David Ing.
\newblock Category {{Number Impacts Rule-Based}} but {{Not Information-Integration Category Learning}}: {{Further Evidence}} for {{Dissociable Category-Learning Systems}}.
\newblock {\em Journal of Experimental Psychology. Learning, Memory \& Cognition}, 30(1):227--245, January 2004.

\bibitem{MilesMatsukiMinda14}
Sarah~J. Miles, Kazunaga Matsuki, and John~Paul Minda.
\newblock Continuous executive function disruption interferes with application of an information integration categorization strategy.
\newblock {\em Attention, Perception, \& Psychophysics}, 76(5):1318--1334, July 2014.

\bibitem{MindaRabi15}
John~P. Minda and Rahel Rabi.
\newblock Ego depletion interferes with rule-defined category learning but not non-rule-defined category learning.
\newblock {\em Frontiers in Psychology}, 6, January 2015.

\bibitem{QuamWangMaddoxEtAl18}
Carolyn Quam, Alisa Wang, W.~Todd Maddox, Kimberly Golisch, and Andrew Lotto.
\newblock Procedural-{{Memory}}, {{Working-Memory}}, and {{Declarative-Memory Skills Are Each Associated With Dimensional Integration}} in {{Sound-Category Learning}}.
\newblock {\em Frontiers in Psychology}, 9, October 2018.

\bibitem{WaldronAshby01}
Elliott~M. Waldron and F.~Gregory Ashby.
\newblock The effects of concurrent task interference on category learning: {{Evidence}} for multiple category learning systems.
\newblock {\em Psychonomic Bulletin \& Review}, 8(1):168--176, 2001.

\bibitem{NosofskyKruschke02}
Robert~M. Nosofsky and John~K. Kruschke.
\newblock Single-system models and interference in category learning: {{Commentary}} on {{Waldron}} and {{Ashby}} (2001).
\newblock {\em Psychonomic Bulletin \& Review}, 9(1):169--174, March 2002.

\bibitem{CasaleRoederAshby12}
Michael~B. Casale, Jessica~L. Roeder, and F.~Gregory Ashby.
\newblock Analogical transfer in perceptual categorization.
\newblock {\em Memory \& Cognition}, 40(3):434--449, April 2012.

\bibitem{GanZhengWangEtAl23}
Zhenzhong Gan, Lurong Zheng, Suiping Wang, and Gangyi Feng.
\newblock Distribution-dependent representations in auditory category learning and generalization.
\newblock {\em Frontiers in Psychology}, 14, September 2023.

\bibitem{Huang-PollockMaddoxKaralunas11}
Cynthia~L. {Huang-Pollock}, W.~Todd Maddox, and Sarah~L. Karalunas.
\newblock Development of implicit and explicit category learning.
\newblock {\em Journal of Experimental Child Psychology}, 109(3):321--335, July 2011.

\bibitem{RabiMilesMinda15}
Rahel Rabi, Sarah~J. Miles, and John~Paul Minda.
\newblock Learning categories via rules and similarity: {{Comparing}} adults and children.
\newblock {\em Journal of Experimental Child Psychology}, 131:149--169, March 2015.

\bibitem{RabiMinda14}
Rahel Rabi and John~Paul Minda.
\newblock Rule-{{Based Category Learning}} in {{Children}}: {{The Role}} of {{Age}} and {{Executive Functioning}}.
\newblock {\em PLOS ONE}, 9(1):e85316, January 2014.

\bibitem{RoarkLeschtHamptonWrayEtAl23}
Casey~L. Roark, Erica Lescht, Amanda Hampton~Wray, and Bharath Chandrasekaran.
\newblock Auditory and visual category learning in children and adults.
\newblock {\em Developmental Psychology}, 59(5):963--975, May 2023.

\bibitem{MindaDesrochesChurch08}
John~Paul Minda, Amy~S. Desroches, and Barbara~A. Church.
\newblock Learning rule-described and non-rule-described categories: A comparison of children and adults.
\newblock {\em Journal of Experimental Psychology. Learning, Memory, and Cognition}, 34(6):1518--1533, November 2008.

\bibitem{CarpenterWillsBenattayallahEtAl16}
Kathryn~L. Carpenter, Andy~J. Wills, Abdelmalek Benattayallah, and Fraser Milton.
\newblock A {{Comparison}} of the neural correlates that underlie rule-based and information-integration category learning.
\newblock {\em Human Brain Mapping}, 37(10):3557--3574, October 2016.

\bibitem{MiltonBealingCarpenterEtAl17}
Fraser Milton, Pippa Bealing, Kathryn~L. Carpenter, Abdelmalek Bennattayallah, and Andy~J. Wills.
\newblock The {{Neural Correlates}} of {{Similarity-}} and {{Rule-based Generalization}}.
\newblock {\em Journal of Cognitive Neuroscience}, 29(1):150--166, January 2017.

\bibitem{BuchsbaumGreerChangEtAl05}
Bradley~R. Buchsbaum, Stephanie Greer, Wei-Li Chang, and Karen~Faith Berman.
\newblock Meta-analysis of neuroimaging studies of the {{Wisconsin Card-Sorting}} task and component processes.
\newblock {\em Human Brain Mapping}, 25(1):35--45, 2005.

\bibitem{Milner63}
Brenda Milner.
\newblock Effects of {{Different Brain Lesions}} on {{Card Sorting}}: {{The Role}} of the {{Frontal Lobes}}.
\newblock {\em Archives of Neurology}, 9(1):90--100, July 1963.

\bibitem{WallisAndersonMiller01}
J.~D. Wallis, K.~C. Anderson, and E.~K. Miller.
\newblock Single neurons in prefrontal cortex encode abstract rules.
\newblock {\em Nature}, 411:953--956, June 2001.

\bibitem{HuntHayden17a}
Laurence~T. Hunt and Benjamin~Y. Hayden.
\newblock A distributed, hierarchical and recurrent framework for reward-based choice.
\newblock {\em Nature reviews. Neuroscience}, 18(3):172--182, February 2017.

\bibitem{CavanaghHuntKennerley20}
Sean~E. Cavanagh, Laurence~T. Hunt, and Steven~W. Kennerley.
\newblock A {{Diversity}} of {{Intrinsic Timescales Underlie Neural Computations}}.
\newblock {\em Frontiers in Neural Circuits}, 14:615626, December 2020.

\bibitem{RussinOReillyBengio20a}
Jacob Russin, Randall~C O'Reilly, and Yoshua Bengio.
\newblock Deep learning needs a prefrontal cortex.
\newblock In {\em Bridging {{AI}} and {{Cognitive Science}} ({{BAICS}}) {{Workshop}}, {{ICLR}} 2020}, page~11, 2020.

\bibitem{TraylorMerulloFrankEtAl24a}
Aaron Traylor, Jack Merullo, Michael~J. Frank, and Ellie Pavlick.
\newblock Transformer {{Mechanisms Mimic Frontostriatal Gating Operations When Trained}} on {{Human Working Memory Tasks}}.
\newblock {\em Proceedings of the Annual Meeting of the Cognitive Science Society}, 46(0), 2024.

\bibitem{SoniTraylorMerulloFrankEtAl24a}
Aneri Soni, Aaron Traylor, Jack Merullo, Michael~J. Frank, and Ellie Pavlick.
\newblock Transformer {{Mechanisms Mimic Frontostriatal Gating Operations When Trained}} on {{Human Working Memory Tasks}}.
\newblock in prep.

\bibitem{KingmaBa15}
Diederik~P. Kingma and Jimmy Ba.
\newblock Adam: {{A Method}} for {{Stochastic Optimization}}.
\newblock In Yoshua Bengio and Yann LeCun, editors, {\em 3rd {{International Conference}} on {{Learning Representations}}, {{ICLR}} 2015, {{San Diego}}, {{CA}}, {{USA}}, {{May}} 7-9, 2015, {{Conference Track Proceedings}}}, 2015.

\end{thebibliography}

% \newpage
\end{document}